\documentclass[sigconf]{acmart}
\usepackage{amsmath,amsfonts}
\usepackage{algorithmic}
\usepackage{graphicx}
\usepackage{algorithm,algorithmic}
\usepackage{hyperref}
\hypersetup{hidelinks=true}
\usepackage{textcomp}
\usepackage{xcolor}
\usepackage[table]{xcolor}
\definecolor{Aqua}{RGB}{0,255,255}
\usepackage{xspace}
\usepackage{multirow}
\usepackage{multicol}
\usepackage{subcaption}
\usepackage{color}
\usepackage{colortbl}
\usepackage{booktabs}
\usepackage{ulem}
\usepackage{dsfont}
\usepackage{enumitem}

\newcommand{\eg}{\textit{e.g.},\xspace}
\newcommand{\modelname}{HDMoE\xspace}
\newcommand{\re}[1]{{\color{black}#1}}
\AtBeginDocument{%
  }

\setcopyright{acmlicensed}
\copyrightyear{2026}
\acmYear{2026}
\acmDOI{XXXXXXX.XXXXXXX}
\acmConference[Conference acronym 'XX]{Make sure to enter the correct
  conference title from your rights confirmation email}{June 03--05,
  2018}{Woodstock, NY}
\acmISBN{978-1-4503-XXXX-X/2018/06}

\acmSubmissionID{505}

\begin{document}

\title{HDMoE: A Hierarchical Decoupling-Fusion Mixture-of-Experts Framework for Multimodal Cancer Survival Prediction}

\author{Huayi Wang}
\email{huayiwang@zju.edu.cn}
\orcid{0009-0003-3371-1482}
\affiliation{%
  \institution{Zhejiang University}
  \city{Hangzhou}
  \country{China}
}

\author{Haochao Ying}
\email{haochaoying@zju.edu.cn}
\orcid{0000-0001-7832-2518}
\authornote{Corresponding authors: Haochao Ying and Ying Sun.}
\affiliation{%
  \institution{Zhejiang University}
  \city{Hangzhou}
  \country{China}}

\author{Yuyang Xu}
\email{xuyuyang@zju.edu.cn}
\affiliation{%
  \institution{Zhejiang University}
  \city{Hangzhou}
  \country{China}
}

\author{Qiyao Zheng}
\email{20222501297@stu.xju.edu.cn}
\affiliation{%
  \institution{Xinjiang University}
  \city{Urumqi}
  \country{China}
}

\author{Jun Wang}
\email{wangjun@hzcu.edu.cn}
\affiliation{%
  \institution{Hangzhou City University}
  \city{Hangzhou}
  \country{China}
}

\author{Cheng Zhang}
\email{zhangcheng@sysucc.org.cn}
\affiliation{%
  \institution{Sun Yat-sen University Cancer Center}
  \city{Guangzhou}
  \country{China}
}

\author{Ying Sun}
\email{sunying@sysucc.org.cn}
\authornotemark[1]
\affiliation{%
  \institution{Sun Yat-sen University Cancer Center}
  \city{Guangzhou}
  \country{China}
}

\author{Jian Wu}
\email{wujian2000@zju.edu.cn}
\affiliation{%
  \institution{Zhejiang University}
  \city{Hangzhou}
  \country{China}}



\begin{abstract}
Multimodal survival prediction, a crucial yet challenging task, demands the integration of multimodal medical data (\eg Whole Slide Images (WSIs) and Genomic Profiles) to achieve accurate prognostic modeling. Given the inherent heterogeneity across modalities, the feature decoupling-fusion paradigm has emerged as a dominant approach. 
However, these methods have the following shortcomings: 
(1) fail to reduce the redundant information of modality features before decoupling, which negatively affects the feature decoupling and fusion effect;
(2) lack the ability to model the fine-grained relationships of the features and capture the local information interactions between intra- and inter-modality features. 
To address these issues, we propose a \underline{H}ierarchical \underline{D}ecoupling-Fusion \underline{M}ixture-\underline{o}f-\underline{E}xperts (HDMoE) framework with two levels of MoE and \underline{R}andom \underline{F}eature \underline{R}eorganization (RFR) modules.
In the first-level MoE, shared experts and routed experts are employed to remove redundant information and extract fine-grained specific features within each modality, while the second-level MoE facilitates fine-grained inter-modality feature decoupling. Besides, we design two RFR modules following each level of MoE to finely fuse intra- and inter-modality features, which can help the model capture more fine-grained relationships between modalities.
Extensive experimental results on our private Liver Cancer (LC) and three TCGA public datasets confirm the effectiveness of our proposed method. 
Codes are available at https://github.com/ZJUMAI/HDMoE.
\end{abstract}





\maketitle

\section{Introduction}
Cancer survival analysis plays a crucial role in understanding patient prognosis and guiding clinical decision-making.
Recent studies have shifted focus from unimodal prediction approaches~\cite{AttnMIL,CLAM,TransMIL,SNN} to leveraging more complex multimodal data, thereby enhancing predictive accuracy and capabilities~\cite{MCAT,CMTAs,MOTCat,xiong2024mome}.
Taking WSIs and Genomic Profiles as an example, WSIs provide spatial organization and morphological features of tumor tissues, while Genomic Profiles reveal molecular-level variations, such as mutations and gene expression changes, making them complementary in nature~\cite{lu2021integrating}. However, the effective fusion of these data poses substantial challenges, particularly due to modality heterogeneity and the complex learning of modality relationships~\cite{li2022hfbsurv,Boehm_2022,Sabah_2021}.


Known studies have attempted to alleviate the above problems using cross-attention~\cite{CMTAs,MCAT,MOTCat}, MoE-based progressive fusion~\cite{xiong2024mome}, and modality decoupling-fusion approaches~\cite{CFDL,zhou2024cohort,PIBD}. However, these methods have not fully considered (1) \textit{redundancy within the modality data} and (2) \textit{fine-grained relationships between the modality features}. Here, redundant information refers to repetitive or highly correlated features within modality representations~\cite{yu2024rcnet,jaume2024modeling}. For instance, WSI patches often exhibit similarities in tissue structures across different regions, while Genomic Profiles within the same pathway may display high correlations. Such redundancy can mislead the model into prioritizing repetitive patterns over discriminative signals, impairing its ability to identify task-relevant features during fusion~\cite{zhou2024multimodal}.
Consequently, effectively reducing redundant information can enhance the model's capacity to capture crucial insights within complex modality relationships. Concerning the second point, although there exists a fine-grained correspondence between specific gene expressions and local regions of WSIs~\cite{zheng2024graph} (\eg the relationship between specific gene mutations and the tumor microenvironment), most works~\cite{CFDL,zhou2024cohort,PIBD} still learn relationships at the level of global features, making it difficult to fully capture local interactions within and across modalities.

To address these challenges, we propose HDMoE, a \underline{H}ierarchical \underline{D}ecoupling-Fusion \underline{M}ixture-\underline{o}f-\underline{E}xperts framework, which employs an interleaved design consisting of two levels of identical Sparse MoE~\cite{hwang2023tutel,riquelme2021scaling} and RFR modules, following a purify-then-decouple strategy. Specifically, modality features are first divided into multiple tokens as input to enable fine-grained feature relationship learning. In the first-level MoE, we introduce a shared expert~\cite{dai2024deepseekmoe} to extract unimodal fine-grained generic features, eliminating the redundant intra-modality features. Additionally, routed experts are employed to extract unimodal specific features at the token level. To help the model capture more local feature relationships, purified and specific features from different modalities are randomly reorganized at a fine-grained level. The second-level MoE is then applied for inter-modality feature decoupling. The reorganized fusion features are first divided into multiple fine-grained tokens, after which fine-grained inter-modality shared and specific decoupled features are extracted by the shared and routed experts, respectively. Finally, the decoupled features are randomly reorganized, which can diversify the fine-grained relationships between decoupled features during fusion. Overall, the MoE with shared expert serves to reduce redundancy in modality features, while the Sparse MoE and RFR provide the model with the ability to explore fine-grained feature relationships both within and across modalities.
Our main contributions are summarized as follows:
\begin{itemize}[left=0em]
\item We target the problem of modality-decoupling-fusion-based survival analysis and propose HDMoE to refine modality features before decoupling and effectively capture fine-grained intra- and inter-modality relationships.
\item We design two RFR modules following each MoE level. These modules achieve feature fusion while assisting the model in capturing more fine-grained feature relationships by feature segmentation and reorganization. 
\item We conduct extensive experiments and ablation studies on multiple datasets for cancer survival analysis, demonstrating that our proposed method achieves state-of-the-art performance.
\end{itemize}
\section{Methodology}
\begin{figure*}
    \centering
    \includegraphics[width=0.9\linewidth]{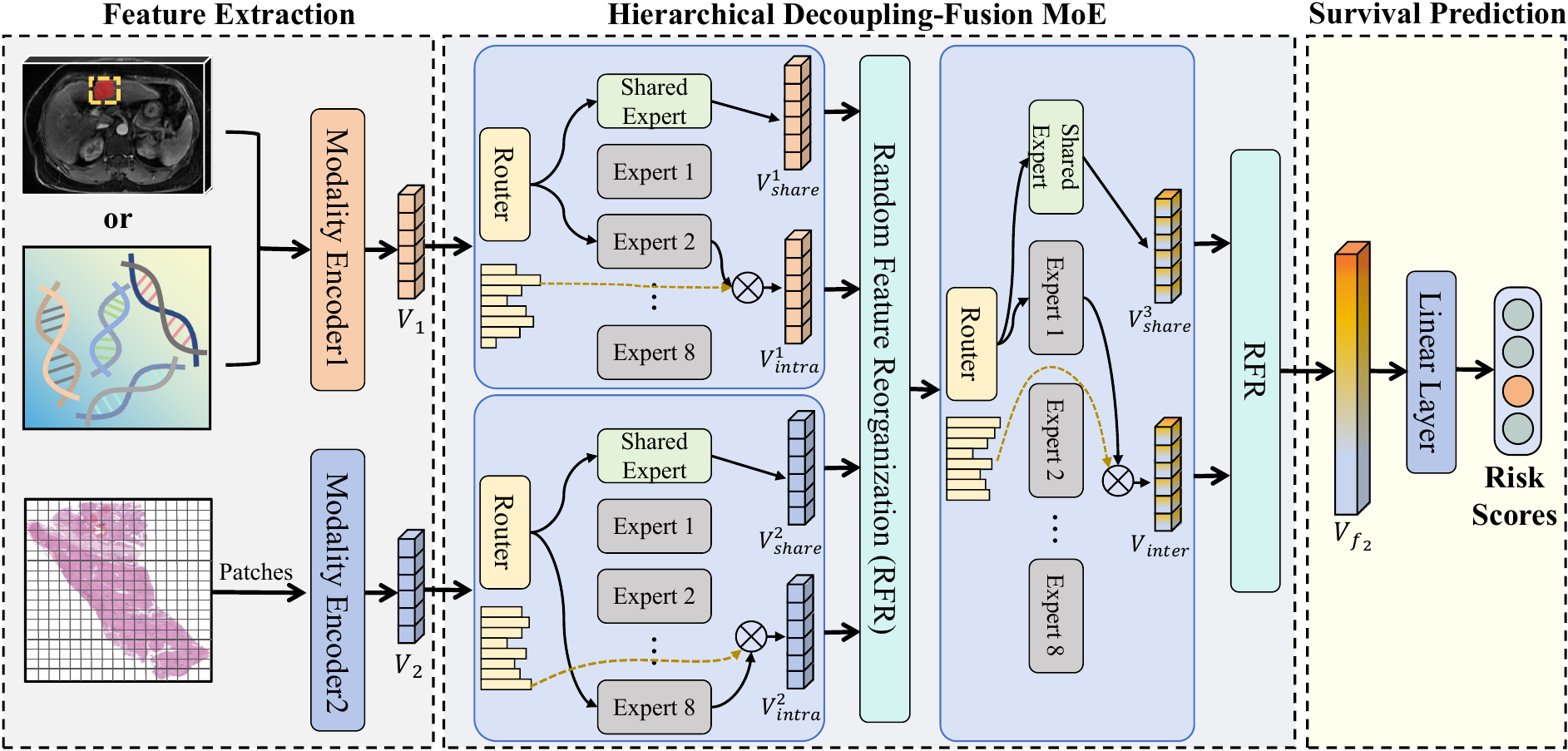}
    \caption{An overview of our proposed framework, consisting of three modules: Feature Extraction, Hierarchical Decoupling-Fusion MoE, and Survival Prediction.}
    \label{fig:framework}
    \vspace{-2ex}
\end{figure*}
\subsection{Overall Framework}
Our framework consists of three main modules: a Feature Extraction module, a Hierarchical Decoupling-Fusion MoE module, and a Survival Prediction module, as shown in Fig.~\ref{fig:framework}.
First, the multimodal data are preprocessed (details are illustrated in Section~\ref{subsec:unimodal}) and fed into corresponding encoders to obtain features $V_1$, $V_2$, respectively. The encoders mainly consist of feature extractor and TransMIL layers~\cite{TransMIL}.
Second, to address the issues of feature redundancy and fine-grained relationship learning in existing modality decoupling-fusion paradigm, we design a Hierarchical Decoupling-Fusion MoE module. The two modalilty features are divided into multiple fine-grained tokens first, and then are fed into the first-level MoE modules to reduce redundant features before decoupling, followed by the RFR module to enhance fine-grained relationship learning. Subsequently, the purified modality features are further processed through the second-level MoE for decoupling, and the fusion features are fused again via RFR module.
Finally, a fully connected layer is used to output the final prediction.

\subsection{Unimodal Feature Extraction}
\label{subsec:unimodal}
Our pipeline initiates by extracting unimodal representations from the patient’s WSIs alongside either MRI scans or Gene Profiles.
Specifically, due to the massive resolution of WSIs, following the common setting~\cite{CMTAs,MCAT,MOTCat}, we adopt the CLAM~\cite{CLAM} to divide tissue regions into non-overlapping $256\times256$ patches at a 20$\times$ magnification level. We then utilize an ImageNet pre-trained ResNet50~\cite{Resnet}
to extract patch features, represented as $P\in \mathbb{R}^{n_p\times d_1}$, where $n_p$ is the number of patches in WSIs and $d_1$ is the dimensionality of the modality latent features. 
%
For MRI scans, we first crop the tumor-relevant regions based on tumor segment labels, and then also employ ResNet50 to extract spatial and sequential features of the tumor. After that, global average pooling is applied to obtain general MRI features, resulting in $M \in \mathbb{R}^{n_m \times d_1}$, where $n_m$ is the number of MRI features. 
For genomic data, we extract genomic sub-sequence features following SNN~\cite{SNN}, represented as $G\in \mathbb{R}^{n_g\times d_1}$, where $n_g$ is the number of genomic sub-sequences. 
After that, the learned modality-aware class tokens ($V_1, V_2 \in \mathbb{R}^{1\times d_1}$ represent $P$ and $M$ or $G$, respectively) derived from TransMIL layers~\cite{TransMIL} are viewed as modality representations, which aim to equalize the number of different modality features. 

\subsection{Dual-Level Mixture of Decoupled Experts}
To enhance the model's capability for fine-grained feature learning, we configure the Sparse MoE as the core component of our framework. 
In fact, the use of Sparse MoE for fine-grained, token-level processing has been well established in prior works~\cite{shazeer2017outrageously,lepikhin2020gshard,fedus2022switch,gross2017hard,Shao2024DeepSeekV2AS}, where different experts are dynamically activated to handle distinct semantic or spatial patterns in individual fine-grained tokens—such as in vision transformers with MoE layers, where each image patch is processed by a sparse subset of experts based on its content.
Building on this, we propose a hierarchical MoE framework that follows a ``purify first, then decouple'' strategy:


\noindent \textbf{First-Level MoE.} In traditional MoE framework, routed experts may redundantly learn common knowledge from tokens. By introducing a shared expert to integrate general knowledge, routed experts can focus on capturing specific information, reducing redundancy and improving efficiency~\cite{Shao2024DeepSeekV2AS}. 
Therefore, we configure shared and routed experts to refine intra-modality features from global and local perspectives, respectively. The shared expert captures the common information across all token features by sharing the parameters between feature tokens (\eg common pathway-level activity patterns in genomics or general tissue architecture semantics in WSIs), reducing redundant information (Experiment results are shown in Fig.~\ref{fig:der}). 
Besides, routed experts extract modality-specific features (\eg mutation-associated semantic patterns in genomics or distinct morphology-related semantic patterns in WSIs)
Specifically, we first divide feature $V_i$ into $T_1$ fine-grained tokens $\{v_t^i\}_{t=1}^{T_1}$ ($i \in \{1,2\}$) by reshaping along the feature dimension, where $v^i_t \in \mathbb{R}^{1\times (d_1/T_1)}$. Given $N+1$ experts $\{E_1^i,...,E_N^i, E_{share}^i\}$ and a router $G_{intra}^i$ that activates the Top-K experts:
\begin{equation}
\begin{gathered}
G_{intra}^i(v_{t}^i) = W_{intra}^i \cdot v_{t}^i \\g_{intra}^i(v_t^i)=\operatorname{Top-K}(\operatorname{Softmax}(G_{intra}^i(v_{t}^i))),\\
V_{intra}^i = \operatorname{Concat}\left(\left\{g_{intra}^i(v_{t}^i) \cdot E_j^{i}(v_{t}^i)\right\}_{t=1}^{T_1}\right) ,\\
V_{share}^i = E_{share}^i(\{v_t^i\}_{t=1}^{T_1}),
\end{gathered}
\label{eq:1moe}
\end{equation}
\noindent where $W_{intra}^i$ is a learnable weight matrix for feature $v_t^i$, $g_{intra}^i$ denotes the routing score, $j$ denotes the selected expert index from $g_{intra}^i$, and $\{V_{intra}^i, V_{share}^i\}\in \mathbb{R}^{1\times d_1}$. 

\noindent \textbf{Second-Level MoE.} The second-level MoE is designed to achieve inter-modality feature decoupling and adopts the same architecture as the First-level MoE.
The shared and routed experts are utilized to capture fine-grained shared features (\eg the correlations between gene mutation patterns and histopathological morphology) and specific features (genomics-specific regulatory signals or pathology-specific morphological cues that are not shared across modalities), for decoupling purposes. 
Specifically, we use $v_{f_{1}}\in\mathbb{R}^{1\times (d_2/T_2)}$ represents one of $T_2$ tokens in $V_{f_1}$ (obtained by RFR, see Eq.~(\ref{eq:rfrm}) below), where $d_2$ and $T_2$ denote the feature dimension and the number of fine-grained tokens in the Second-Level MoE, respectively. Given another $N+1$ experts $\{E_1^{inter},...,E_N^{inter}, E_{share}^3\}$ and a router $G_{inter}$: 
\begin{equation}
\begin{gathered}
G_{inter}(v_{f_1}) = W_{inter} \cdot v_{f_1},\\
g_{inter}(v_{f_1})=\operatorname{Top-K}(\operatorname{Softmax}(G_{inter}(V_{f_1}))), \\
V_{inter} = \operatorname{Concat}\left(\left\{g_{inter}(v_{f_1}) \cdot E_j^{inter}(v_{f_1})\right\}_1^{T_2}\right),\\
V_{share}^{3} = E_{share}^3(\{v_{f_1}\}_{1}^{T_2}),
\end{gathered}
\label{eq:2moe}
\end{equation}
where $W_{inter}$ is a learnable weight matrix for feature $v_{f1}$, $g_{inter}$ denotes the routing score, and $j$ denotes the selected expert index from $g_{inter}$, and $\{V_{inter}, V^3_{share}\}\in \mathbb{R}^{1\times d_2}$. 

\begin{figure}
    \centering
    \includegraphics[width=0.6\linewidth]{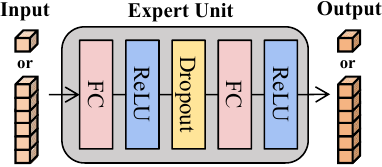}
    \caption{An overview of Expert Unit Framework.}
    \label{fig:exp}
    \vspace{-2ex}
\end{figure}
Furthermore, all expert units are composed of the same feed-forward network framework, as shown in Figure~\ref{fig:exp}. Using the modality feature $V_1$ as an example, the shared expert is responsible for capturing the common information across all tokens $\{v_t^1\}_{t=1}^{T_1}$, while the routed experts focus on extracting fine-grained modality-specific information from token $v_t^1$.
 
\subsection{Random Feature Reorganization (RFR)}
\begin{figure}
    \centering
    \includegraphics[width=\linewidth]{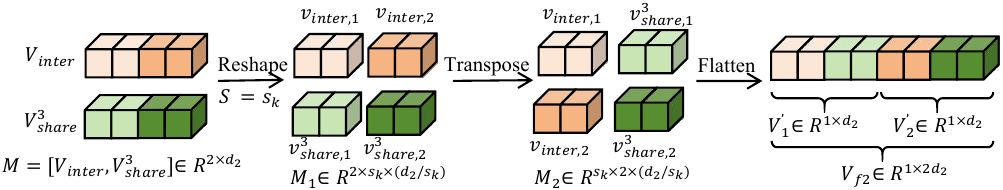}
    \caption{Illustration of Random Feature Reorganization.}
    \label{fig:rfr}
    \vspace{-2ex}
\end{figure}

To facilitate the model in capturing more information about the relationships among fine-grained features, we design a RFR module. This module is capable of achieving feature fusion while further diversifying the combinations of fine-grained features. The process can be formulated as Eq. (\ref{eq:rfrm}),
\begin{equation}
\begin{gathered}
V_{f_1}=\operatorname{RFR}_1(\{V_{intra}^1,V_{share}^1,V_{intra}^2,V_{share}^2\} ),\\
V_{f_2}=\operatorname{RFR}_2(\{V_{inter},V_{share}^3\}),
\end{gathered}
\label{eq:rfrm}
\end{equation}
\noindent where $\operatorname{RFR}_1$ is the first-level RFR following the first-level MoE, positioned to enhance the second-level MoE's capacity for capturing fine-grained local features, while the $\operatorname{RFR}_2$ is deployed downstream of the second-level MoE, specifically designed to optimize feature diversification by establishing heterogeneous relational patterns among the refined feature representations.
Specifically, we first set $n$ segment values $S=\{s_k\}_{k=1}^{n}$ to divide each MoE output $V_o$ into $L$ equal feature segments $\{v_{o,l}\}_{l=1}^L$, where 
$V_o$ represents any one of $\{V_{intra}^1,V_{share}^1,V_{intra}^2,V_{share}^2,V_{inter},V_{share}^3\}$. After that, each feature segments is reorganized. Taking $V_{f_2}$ as an example, the process can be formulated as Eq.~(\ref{eq:rfrm_2}),
\begin{equation}
\begin{gathered}
V_{f_2}=\operatorname{Concat}(\{V_{l}^{'}\}_{l=1}^L), \\
V_{l}^{'}=\operatorname{Concat}(\{v_{inter,l},v_{share,l}^3\} ).
\end{gathered}
\label{eq:rfrm_2}
\end{equation}
Furthermore, we clearly illustrate the implementation of RFR in Fig.~\ref{fig:rfr}. This algorithm can be elegantly implemented through matrix transformations. Also, use $V_{f_2}$ as an example, the intermediate matrices are denoted as $M, M_1$, and $M_2$. First, $V_{inter}$ and $V^3_{share}$ are concatenated to form $M=[V_{inter},V^3_{share}]\in \mathbb{R}^{2\times d_2}$. Second, $M$ is reshaped according to the segment value $s_k$ to obatin $M_1\in \mathbb{R}^{2\times s_k\times(d_2/s_k)}$. Third, $M_1$ is transposed along its first and second dimensions to yield $M_2\in \mathbb{R}^{s_k\times 2\times (d_2/s_k)}$. Finally, flattening $M_2$ can directly produces the reorganized feature $V_{f_2}\in \mathbb{R}^{1\times 2d_2}$.
Notably, the segment value is randomly chosen at every forward step. 

Overall, RFR is a novel feature fusion paradigm, whose advantages lie in: (1) compared to direct concatenation, it enables diversified feature combinations, mitigates model’s overfitting to fixed feature combinations, and forces generalization to broader interaction patterns; (2) explicitly constructs fine-grained feature relationships, compelling the model to discover cross-modal subspace correlations.
Moreover, this is similar to PatchShuffle~\cite{kang2017patchshuffle} in image processing—disrupting feature arrangements paradoxically enhances the model's learning of spatial invariance.
\subsection{Survival Prediction}
In survival prediction module, we use one fully connected layer to estimate the hazard function $h(t)=h(T=t \mid T \geq t,(V_1, V_2)) \in[0,1]$, which is the probability of death for a patient right after the time point $t$. 
To achieve decoupling, we employ a Distance Metric (DM) function to control the relationships between different features. 
By default, we use cosine similarity as DM function.
Specifically, features within each MoE level are kept apart, while features across MoE level are kept closer, as shown in Eq. (\ref{eq:loss_dis}). Functions $\text{DM}$ and $\text{DM} ^{'}$ are inversely related to each other.
\begin{equation}
   \begin{split}
   \mathcal{L}_{\text{dm}}&=\sum_{i=1}^2 \operatorname{DM}^{'}(V_{intra}^i,V_{share}^i)+\operatorname{DM}^{'}(V_{inter},V_{share}^3)\\&+
   \operatorname{DM}(\{V_{intra}^i\}_{i=1}^2,V_{inter}) +\operatorname{DM}(\{V_{share}^i\}_{i=1}^2,V_{share}^3).
    \end{split}
    \label{eq:loss_dis}
\end{equation}
Additionally, to prevent the majority of tokens from being routed to a small subset of experts, which could lead to insufficient training of the remaining experts~\cite{shazeer2017sparsely}, we introduce a load balance loss~\cite{lepikhin2020gshard,wei2024skywork} as follows:
\begin{equation}
\begin{gathered}
    \mathcal{L}_{\text{bl}}=\sum_{i=1}^N f_i P_i, f_i=\frac{1}{T} \sum_{t=1}^T \mathds{1}(v_t \text{ selects Expert } j), \\
    P_i=\frac{1}{T} \sum_{t=1}^T \operatorname{Softmax}(W \cdot v_t),
\end{gathered}
    \label{eq:bl_loss}
\end{equation}
where $T$ denotes the number of tokens, $\mathds{1}(*)$ is the indicator function, and $W$ represents learnable weight from gating network. The total loss is shown in Eq.~(\ref{eq:totalloss}), 
\begin{equation}
    \mathcal{L} = \mathcal{L}_{\text {surv }}+ \alpha \mathcal{L}_{\text{dm}} + \beta \mathcal{L}_{\text{bl}},
    \label{eq:totalloss}
\end{equation}
where $\mathcal{L}_{\text {surv}}$ is the survival prediction loss (Appendix \ref{ap:loss}), $\alpha$ and $\beta$ are balance factors.

\section{Experiments and Results}
\subsection{Datasets}
We conduct extensive experiments on a private LC dataset as well as three public TCGA\footnote{https://portal.gdc.cancer.gov} (The Cancer Genome Atlas) datasets to evaluate the performance.
The LC Dataset comprises 160 pairs of diagnostic MRIs and WSIs annotated with ground-truth survival outcomes. Each MRIs cover five distinct sequences per patient: Arterial phase, Delay phase, Hepatobiliary phase, T1-weighted imaging, T2-weighted imaging, while the WSIs are stained using H\&E method~\cite{lillie1954histopathologic}.
The TCGA datasets include paired WSIs and genomic data. We use three typical cancer types: Bladder Urothelial Carcinoma (BLCA, n=373), Breast Invasive Carcinoma (BRCA, n=956), and Lung Adenocarcinoma (LUAD, n=453). Consistent with prior studies~\cite{MCAT}, we categorize the genomic data into six sequences: Tumor Suppression, Oncogenesis,  Protein Kinases, Cellular Differentiation, Transcription, and Cytokines and Growth.

\subsection{Implementation Details}
\textbf{Training Settings.} We compare HDMoE with unimodal and multimodal baselines, including: SNN~\cite{SNN}, SNNTrans~\cite{SNN}, MaxMIL~\cite{wang2018revisiting}, MeanMIL~\cite{zaheer2017deep}, AttnMIL~\cite{AttnMIL}, CLAM~\cite{CLAM}, TransMIL~\cite{TransMIL}, MCAT~\cite{MCAT}, CMTA~\cite{CMTAs}, MOTCAT~\cite{MOTCat}, CFDL~\cite{CFDL}, PIBD~\cite{PIBD}, CCL~\cite{zhou2024cohort},  MoME~\cite{xiong2024mome} and LD-CVAE~\cite{LD-CVAE}. For LC dataset, ResNet3D~\cite{Resnet} is used as MRIs-related unimodal baseline model. The C-Index~\cite{harrell1982evaluating} is adopted as the evaluation metric.
In our experiments, we perform 5-fold cross-validation.

\noindent \textbf{Hyper-parameters.} For the proposed HDMoE, we implement the framework by setting $d_1 = 256$, $d_2 = 512$, $n_m = 1$, and $n_g = 6$. 
We use Adam optimizer with a learning rate of $5\times10^{-4}$ for all datasets. 
The batchsize is set to 1, weight decay is set to $1\times10^{-3}$, and the number of epochs is 30. After optimizing the hyper-parameters, we set the balance factors $\alpha=1$, $\beta=0.01$, numbers of fine-grained tokens $T_1=4$, $T_2=16$, numbers of experts $N=8$, Top-K $=1$ and segment values $S=\{1,2,4,8,16,32,64,128\}$.
\begin{table}[t]
    \caption{C-Index (mean ± std) performance over LC dataset. Img. and Patho. refer to Imaging modality (MRIs) and Pathology modality (WSIs).
    The best results are underlined in \textcolor{red}{\underline{red}}, while the second best are italicized in \textcolor{blue}{\textit{blue}}, respectively.}
\label{tab:lc}
    \centering
    \renewcommand{\arraystretch}{0.85} 
    \begin{tabular}{l|cc|c}
        \toprule
        \multirow{2}{*}{\bfseries Model} &\multicolumn{2}{c|}{\bfseries Modality}
         &\multirow{2}{*}{\bfseries C-index }
         \\
         &  Img. & Patho. \\
        \midrule
        ResNet3D \cite{Resnet}  & \checkmark   &  &
         0.645$\pm$0.036  \\
        \midrule
        AttnMIL \cite{AttnMIL} & &\checkmark &0.520$\pm$0.052 \\
        CLAM\_MB \cite{CLAM}   &    & \checkmark & 0.510$\pm$0.053  \\
        CLAM\_SB \cite{CLAM}   &    & \checkmark  & 0.518$\pm$0.041 \\
        TransMIL \cite{TransMIL}   &    & \checkmark  & 0.592$\pm$0.047 \\
        \midrule
        MCAT \cite{MCAT}   & \checkmark   & \checkmark  & 0.614$\pm$0.040 \\
        CMTA \cite{CMTAs}   & \checkmark   & \checkmark  & 0.597$\pm$0.079 \\
        MOTCat \cite{MOTCat}   & \checkmark   & \checkmark  & 0.621$\pm$0.043 \\
        CFDL \cite{CFDL}   & \checkmark   & \checkmark  & 0.612$\pm$0.053 \\
        CCL \cite{zhou2024cohort} & \checkmark   & \checkmark  & 0.648$\pm$0.034
         \\
        MoME \cite{xiong2024mome}   & \checkmark   & \checkmark  & \textcolor{blue}{\textit{0.650}}$\pm$0.037 \\
        LD-CVAE \cite{LD-CVAE} & \checkmark   & \checkmark  & 0.645$\pm$0.044
         \\
        \rowcolor{Aqua!20} HDMoE (ours)   & \checkmark   & \checkmark  & \textcolor{red}{\underline{0.683}}$\pm$0.029 \\
        \bottomrule
    \end{tabular}
    \vspace{-2ex}
\end{table}

\begin{table*}[t]
    \centering
    \renewcommand{\arraystretch}{0.85} 
    \caption{C-Index (mean ± std) performance over three TCGA datasets. Genomic and Pathology refer to Genomic modality (Genomic profiles) and Pathology modality (WSIs), respectively. The best results are underlined in \textcolor{red}{\underline{red}}, while the second best are italicized in \textcolor{blue}{\textit{blue}}, respectively. 
    }
    \label{tab:tcga}
    \begin{tabular}{l|cc|cccc}
        \toprule        
        \multirow{2}{*}{\bfseries Model} &\multicolumn{2}{c|}{\bfseries Modality}  
        & \multicolumn{4}{c}{\bfseries Dataset} \\
        & Geno. &Patho.
        & BLCA &  BRCA & LUAD & Overall  \\
        
        \midrule
        SNN \cite{SNN}   & \checkmark   &  & 0.618$\pm$0.022  &0.633$\pm$0.074 &0.611$\pm$0.047 & 0.621\\
        SNNTrans \cite{SNN}   & \checkmark  &  &0.659$\pm$0.032  &0.648$\pm$0.058 &0.638$\pm$0.022 & 0.648\\
        \midrule
        MaxMIL~\cite{wang2018revisiting} &  &\checkmark &0.551$\pm$0.032 &0.600$\pm$0.055 &0.596$\pm$0.060 &0.582\\
        MeanMIL~\cite{zaheer2017deep}   &    & \checkmark & 0.585$\pm$0.032 &0.611$\pm$0.029 &0.576$\pm$0.054 & 0.591\\
        AttnMIL \cite{AttnMIL} &  &\checkmark &0.599$\pm$0.048 &0.590$\pm$0.047 &0.620$\pm$0.061 &0.603\\
        CLAM\_MB \cite{CLAM}   &    & \checkmark & 0.565$\pm$0.027 &0.620$\pm$0.052 &0.582$\pm$0.072 & 0.589\\
        CLAM\_SB \cite{CLAM}   &    & \checkmark  & 0.559$\pm$0.034 &0.609$\pm$0.033 &0.594$\pm$0.063 &0.587\\
        TransMIL \cite{TransMIL}   &    & \checkmark  & 0.575$\pm$0.034 &0.643$\pm$0.037 &0.642$\pm$0.046 &0.620\\
        \midrule
        MCAT \cite{MCAT}   & \checkmark   & \checkmark   &0.672$\pm$0.032 &0.659$\pm$0.042 &0.659$\pm$0.027 &0.663\\
        CMTA \cite{CMTAs}   & \checkmark   & \checkmark   &0.660$\pm$0.026 &0.652$\pm$0.058 &0.669$\pm$0.033 &0.660\\
        MOTCat \cite{MOTCat}   & \checkmark   & \checkmark  & 0.673$\pm$0.037 &0.673$\pm$0.032 &0.661$\pm$0.039 &0.669\\
        CFDL \cite{CFDL}   & \checkmark   & \checkmark  & 0.675$\pm$0.025 &0.658$\pm$0.038 & 0.664$\pm$0.030& 0.666\\
        PIBD \cite{PIBD}   & \checkmark   & \checkmark  & 0.599$\pm$0.039 &0.643$\pm$0.029 &-- &0.621\\
        CCL \cite{zhou2024cohort} & \checkmark   & \checkmark  &0.640$\pm$0.041 &0.669$\pm$0.043 &0.657$\pm$0.037 & 0.655\\
        MoME \cite{xiong2024mome} & \checkmark   & \checkmark  &\textcolor{blue}{\textit{0.674}}$\pm$0.020 &\textcolor{blue}{\textit{0.679}}$\pm$0.035 &\textcolor{blue}{\textit{0.672}}$\pm$0.056 &\textcolor{blue}{\textit{0.675}} \\
        LD-CVAE \cite{LD-CVAE} & \checkmark   & \checkmark  & 0.652$\pm$0.020 &0.663$\pm$0.044 &0.665$\pm$0.027 &0.660\\
         \rowcolor{Aqua!20} HDMoE (Ours)   & \checkmark   & \checkmark  & \textcolor{red}{\underline{0.694}}$\pm$0.019 & \textcolor{red}{\underline{0.686}}$\pm$0.022& \textcolor{red}{\underline{0.675}}$\pm$0.014&\textcolor{red}{\underline{0.685}}\\
        \bottomrule
    \end{tabular}
    \vspace{-2ex}
\end{table*}
\subsection{Comparison Results}
On LC Dataset, results show that our method outperforms all others, see Tab.~\ref{tab:lc}. 
For example, in CCL method ($\downarrow 3.5\%$),
redundant information is not removed before feature decoupling, and the MoME method ($\downarrow 3.3\%$) fails to fully consider the relationship between the fine-grained modality features, and instead uses the overall features as inputs for relationship learning.
These results demonstrate that our proposed \modelname framework is effective. The first-level MoE successfully reduces redundant information from MRIs and WSIs, while the Sparse MoE and RFR modules enhance the model’s capability to learn fine-grained relationships between modality features. 
Furthermore, we observe that the MRI-based method outperforms WSI-based methods. We hypothesize that MRIs, as a temporal and volumetric imaging modality, provide a three-dimensional representation of tumors, crucial for understanding the spatial distribution and morphological characteristics of lesions. In contrast, factors such as staining variations and regional heterogeneity in WSIs may negatively impact image quality and hinder model performance.

Results on three TCGA Datasets are shown in Tab.~\ref{tab:tcga}. 
It can be observed that our method achieves the best performance compared to all unimodal methods. Quantitatively, our method outperforms the best unimodal method by 3.5\%, 3.8\%, and 3.7\%, respectively. Also, our method demonstrates outstanding performance compared to multimodal methods, especially on BLCA and BRCA datasets.
Specifically, compared to MoME method, our method achieves improvements of 2.0\%, 0.7\%, and 0.3\% on three datasets and a 1\% improvement in overall performance. These results indicate that our method is also well-suited for integrating genomic and pathology data. Notably, genomic-based methods generally outperform WSI-based methods and even some multimodal methods, highlighting the critical role of genomic modality and the importance of effectively integrating genomic and pathology data. By hierarchical MoE and RFR modules to eliminate modality redundancies and enhance fine-grained inter-modality interactions, our method improves the efficacy of feature decoupling-fusion paradigm.
\begin{table*}
    \caption{Ablation study on four datasets to evaluate the effectiveness of Random Feature Reorganization (RFR.), Balance loss ($\mathcal{L}_{\text{bl}}$), Decouple loss ($\mathcal{L}_{\text{dm}}$), Fine-grained feature (Ff.), and De-redundancy (Dr.).}
    
\label{tab:ablation}
    \centering
    \renewcommand{\arraystretch}{0.85} 
    \begin{tabular}{l|ccccc|cccc|c}
        \toprule
        \multirow{2}{*}{\bfseries Model} 
        &\multicolumn{5}{c|}{\bfseries Settings}
        &\multicolumn{4}{c|}{\bfseries Dataset}
        &\multirow{2}{*}{\bfseries Overall}
         \\
         & RFR. & $\mathcal{L}_{\text{bl}}$ &$\mathcal{L}_{\text{dm}}$ &Ff. &Dr. &LC &BLCA &BRCA &LUAD &  \\
        \midrule
        Variant 1   & \checkmark   &  & & \checkmark &\checkmark
        & 0.654$_{\pm0.012}$ & 0.665$_{\pm 0.026}$  & 0.660$_{\pm0.058}$ & 0.664$_{\pm 0.023}$ & 0.661\\
        Variant 2   & \checkmark   & \checkmark  & & \checkmark &\checkmark
        & 0.657$_{\pm0.016}$ & 0.670$_{\pm0.036}$ & 0.651$_{\pm0.021}$ & 0.667$_{\pm0.041}$
        & 0.661\\
        Variant 3   &    & \checkmark &\checkmark & \checkmark &\checkmark
        & 0.640$_{\pm0.036}$ & 0.670$_{\pm0.022}$ & 0.661$_{\pm0.020}$ & 0.672$_{\pm0.037}$ & 0.661\\
        Variant 4   & \checkmark   &  &\checkmark & \checkmark &\checkmark
        & \textcolor{blue}{\textit{0.659}}$_{\pm0.019}$ & \textcolor{blue}{\textit{0.681}}$_{\pm0.011}$ & \textcolor{blue}{\textit{0.674}}$_{\pm0.036}$ & 0.665$_{\pm0.032}$ & \textcolor{blue}{\textit{0.670}}\\
        
        Variant 5   & \checkmark   &\checkmark  &\checkmark &  &\checkmark
        & 0.653$_{\pm0.025}$ & 0.666$_{\pm0.025}$ & 0.661$_{\pm0.014}$ & \textcolor{blue}{\textit{0.673}}$_{\pm0.034}$ & 0.663\\
        
        Variant 6   & \checkmark   &\checkmark  &\checkmark &\checkmark  &
        & 0.655$_{\pm0.026}$ & 0.670$_{\pm0.028}$ & 0.645$_{\pm0.021}$ & 0.652$_{\pm0.021}$ & 0.660\\
        \midrule
        
        \rowcolor{Aqua!20} HDMoE   & \checkmark   &\checkmark  &\checkmark & \checkmark &\checkmark
        & \textcolor{red}{\underline{0.683}}$_{\pm0.029}$ & \textcolor{red}{\underline{0.694}}$_{\pm0.019}$ & \textcolor{red}{\underline{0.686}}$_{\pm0.022}$ & \textcolor{red}{\underline{0.675}}$_{\pm0.014}$ & \textcolor{red}{\underline{0.685}}\\
        \bottomrule
    \end{tabular}
\end{table*}

\subsection{Ablation Studies}

\begin{table}
\setlength{\textfloatsep}{1pt} 
 \caption{The impact of different Distance Metrics (DM), including L1 distance (L1), Kullback-Leibler divergence (KL), Mean Square Error (MSE) and Cosine similarity (COS).}
    \centering
    \renewcommand{\arraystretch}{0.85} 
    \begin{tabular}{l|cccc}
    \toprule
        \multirow{2}{*}{\bfseries DM}
        &\multicolumn{4}{c}{\bfseries Dataset}
        \\
     &LC &BLCA &BRCA &LUAD \\
     \midrule
     L1 & 0.647$_{\pm0.027}$ & 0.659$_{\pm0.048}$ & 0.618$_{\pm0.033}$ & \textcolor{blue}{\textit{0.670}}$_{\pm0.035}$\\
     KL & 0.652$_{\pm0.031}$ & 0.648$_{\pm0.013}$ & 0.646$_{\pm0.026}$ & 0.669$_{\pm0.031}$ \\
     MSE& \textcolor{blue}{\textit{0.680}}$_{\pm0.042}$ & 0.644$_{\pm0.048}$ & \textcolor{blue}{\textit{0.648}}$_{\pm0.038}$ & 0.661$_{\pm0.021}$ \\
     COS & \textcolor{red}{\underline{0.683}}$_{\pm0.029}$ 
        & \textcolor{red}{\underline{0.694}}$_{\pm0.019}$
        & \textcolor{red}{\underline{0.686}}$_{\pm0.022}$ 
        & \textcolor{red}{\underline{0.675}}$_{\pm0.014}$\\
    \bottomrule
    \end{tabular}
    \label{tab:dis}
\end{table}

\begin{table}[t]
\centering
\renewcommand{\arraystretch}{0.85} 
\caption{Modality Importance Experiments on four datasets. G./M. represents Genomic or MRI modality, Patho. represents Pathology modality.}
\begin{tabular}{l|cccc}
\toprule
    \multirow{2}{*}{\bfseries Modal}
    &\multicolumn{4}{c}{\bfseries Dataset} \\
  & LC & BLCA & BRCA & LUAD \\
\midrule
G./M. &\textcolor{red}{\underline{0.653}}$_{\pm0.014}$ & \textcolor{red}{\underline{0.664}}$_{\pm0.025}$ & \textcolor{red}{\underline{0.658}}$_{\pm0.021}$ & \textcolor{red}{\underline{0.662}}$_{\pm0.038}$ \\
Patho. &0.634$_{\pm0.028}$& 0.607$_{\pm0.024}$ & 0.626$_{\pm0.014}$ & 0.621$_{\pm0.016}$ \\
\bottomrule
\end{tabular}
\label{tab:modality_performance}
\end{table}


\begin{figure*}[t]
    \centering
    \begin{subfigure}{0.24\textwidth}
        \centering
        \includegraphics[width=0.95\textwidth]{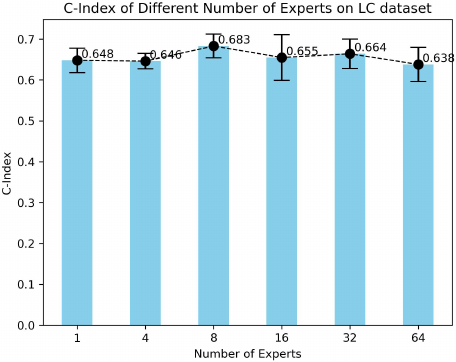}
        \caption{LC dataset}
    \end{subfigure}
    \begin{subfigure}{0.24\textwidth}
        \centering
        \includegraphics[width=0.95\textwidth]{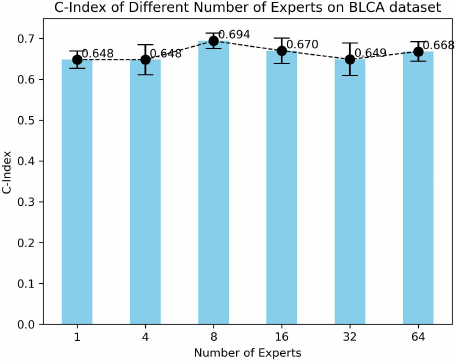}
        \caption{BLCA dataset}
    \end{subfigure}
    \begin{subfigure}{0.24\textwidth}
		\centering
		\includegraphics[width=0.95\textwidth]{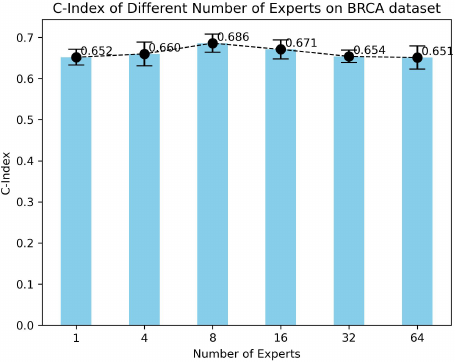}
		\caption{BRCA dataset}
	\end{subfigure}
	\begin{subfigure}{0.24\textwidth}
		\centering
		\includegraphics[width=0.95\textwidth]{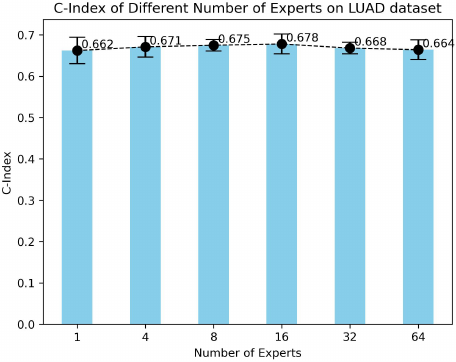}
		\caption{LUAD dataset}
    \end{subfigure}
      \caption{Ablation results of different number of experts on four datasets. 
      }
    \label{fig:num_expert}
        \vspace{-2ex}
\end{figure*}
\begin{figure*}[t]
    \centering
    \begin{subfigure}[t]{0.49\textwidth}
        \centering
        \includegraphics[width=\textwidth]{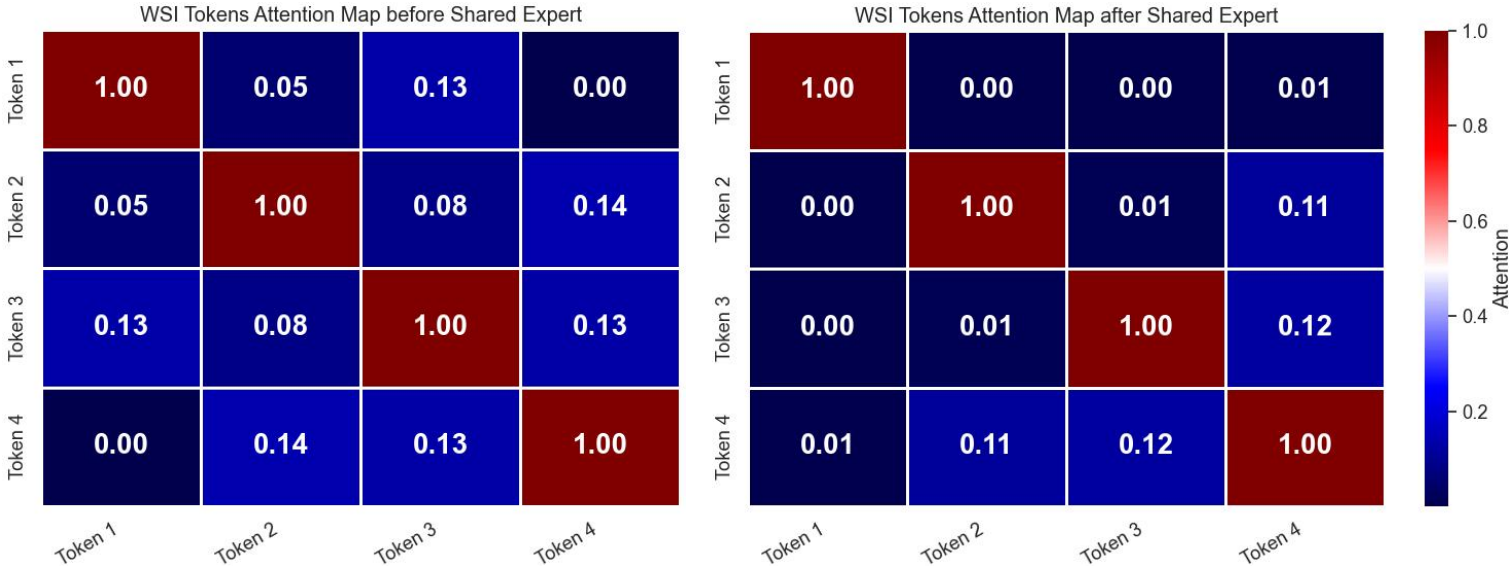}
        \caption{Pathology token correlations before and after shared expert}
    \end{subfigure}
    \hfill
    \begin{subfigure}[t]{0.49\textwidth}
        \centering
        \includegraphics[width=\textwidth]{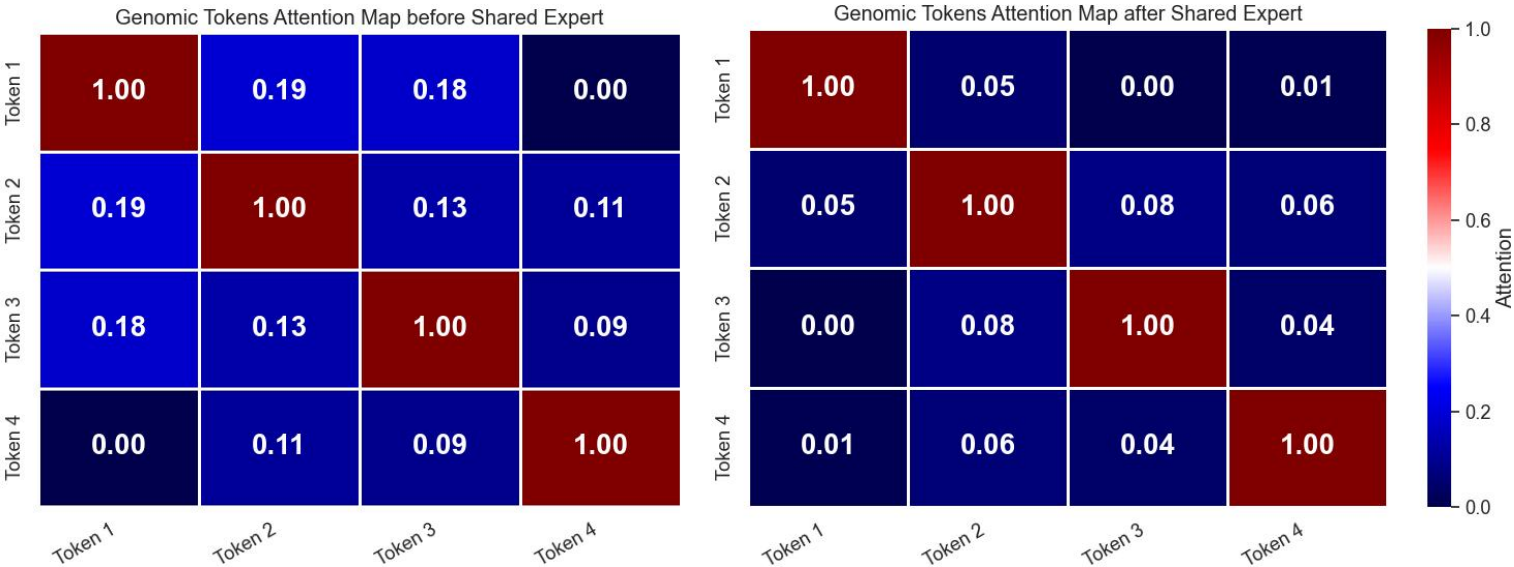}
        \caption{Genomic token correlations before and after shared expert}
    \end{subfigure}
    \caption{Feature de-redundancy experiments on TCGA-BLCA dataset. Each sub-figure shows average correlation heatmaps of tokens before and after shared expert.}
    \label{fig:der}
    \vspace{-2ex}
\end{figure*}
\begin{figure*}[t]
    \centering
    \begin{subfigure}[t]{0.49\textwidth}
        \centering
        \includegraphics[width=\textwidth]{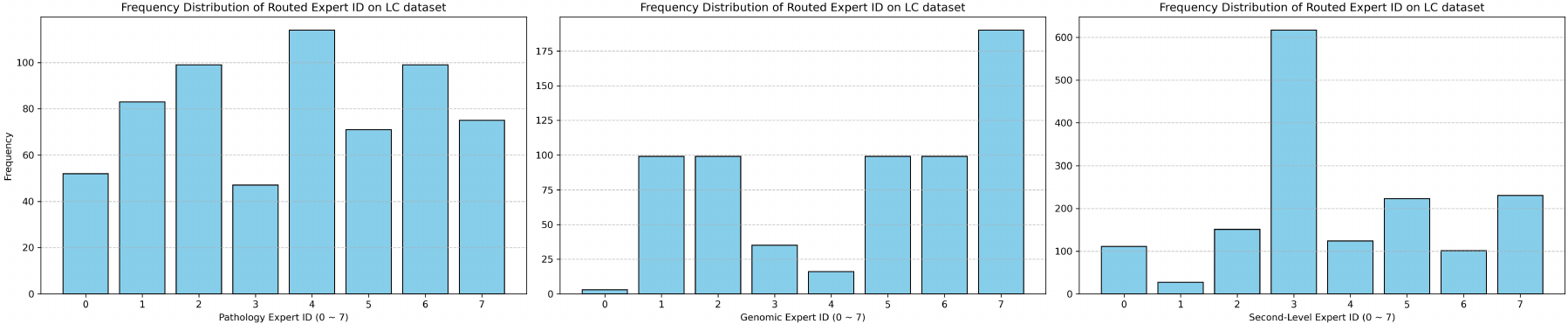}
        \caption{LC dataset}
    \end{subfigure}
    \begin{subfigure}[t]{0.49\textwidth}
        \centering
        \includegraphics[width=\textwidth]{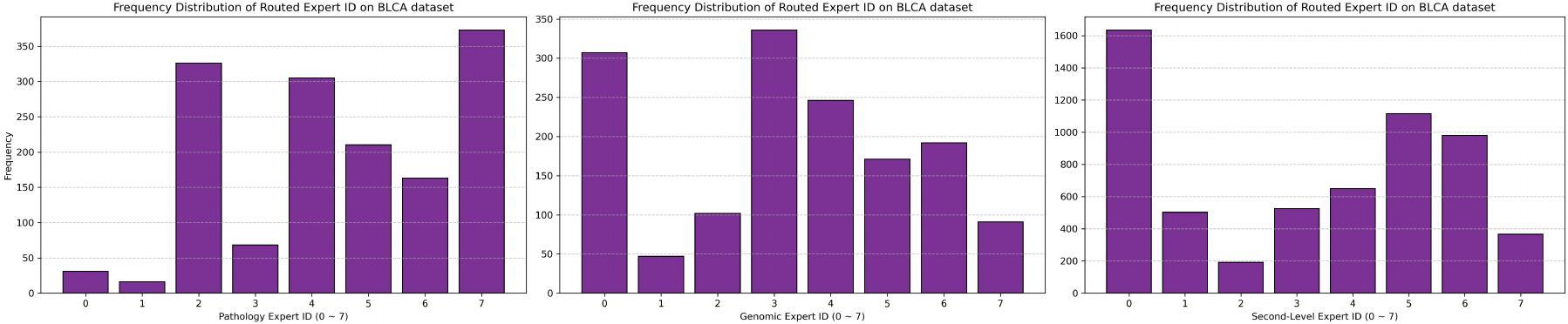}
        \caption{BLCA dataset}
    \end{subfigure}
    \begin{subfigure}[t]{0.49\textwidth}
        \centering
        \includegraphics[width=\textwidth]{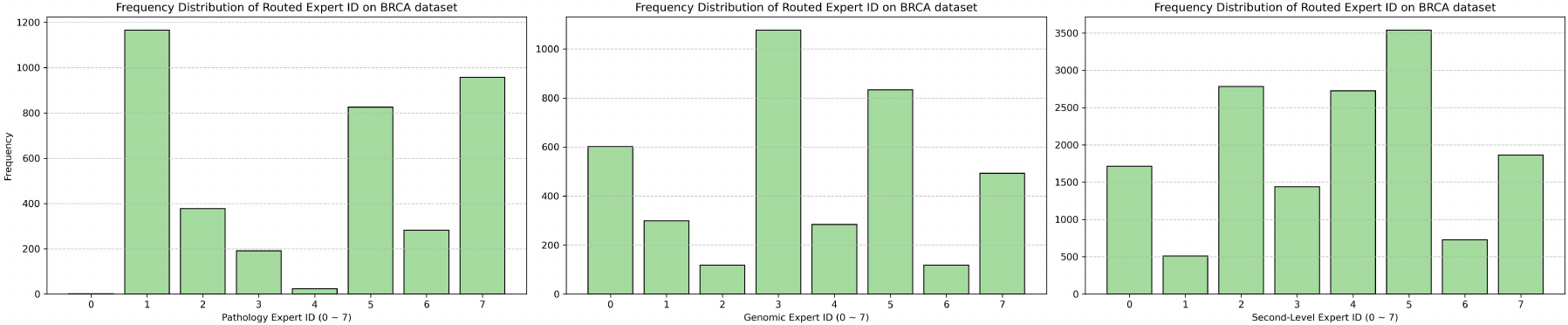}
        \caption{BRCA dataset}
    \end{subfigure}
    \begin{subfigure}[t]{0.49\textwidth}
        \centering
        \includegraphics[width=\textwidth]{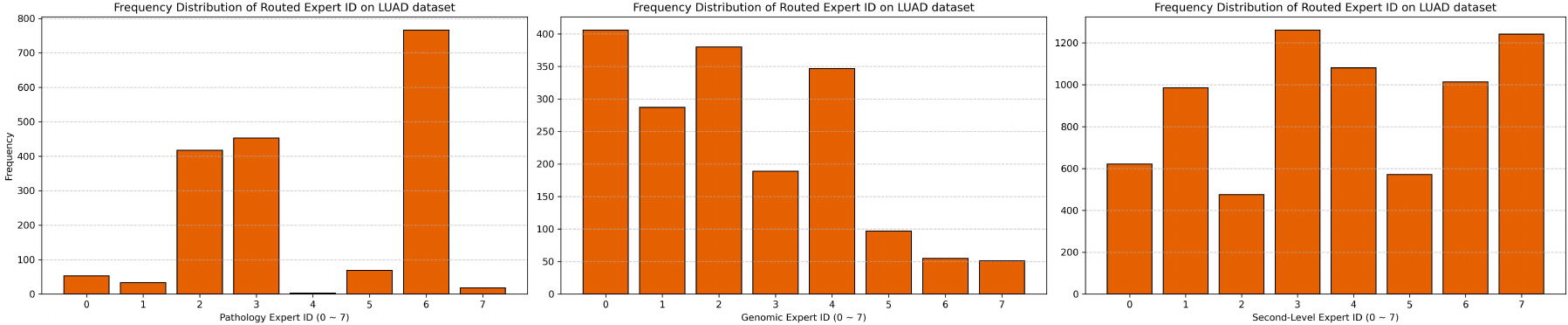}
        \caption{LUAD dataset}
    \end{subfigure}
    \caption{Histograms of routed expert allocations on four datasets. In each sub-figure of the dataset, the left, middle, and right figures respectively show the allocations of First-level pathology experts, First-level genomic experts, and Second-level experts.
    }
    \vspace{-2ex}
    \label{fig:exp_routed}
\end{figure*}

\begin{figure*}[t]
    \centering
    \begin{subfigure}{0.24\textwidth}
        \centering
        \includegraphics[width=0.95\textwidth]{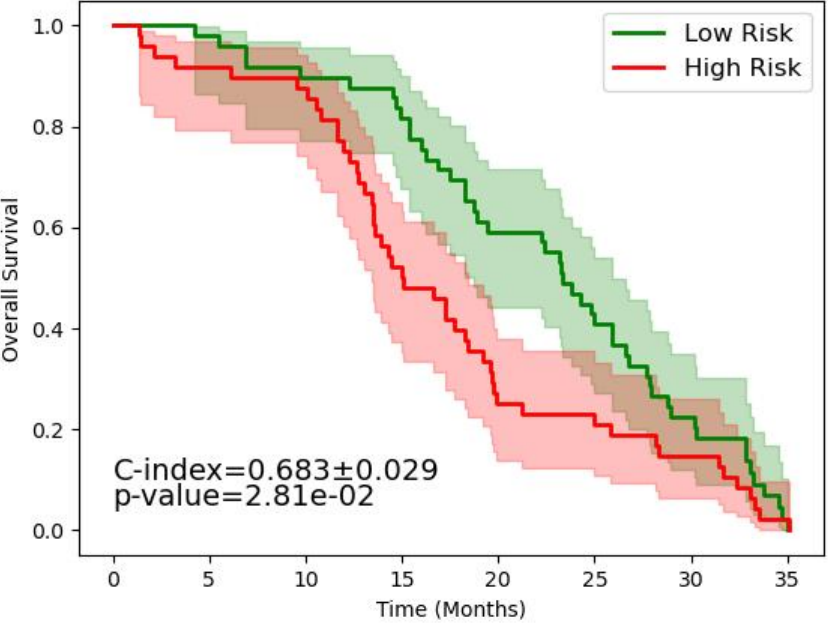}
        \caption{LC dataset}
    \end{subfigure}
    \hfill
    \begin{subfigure}{0.24\textwidth}
        \centering
        \includegraphics[width=0.95\textwidth]{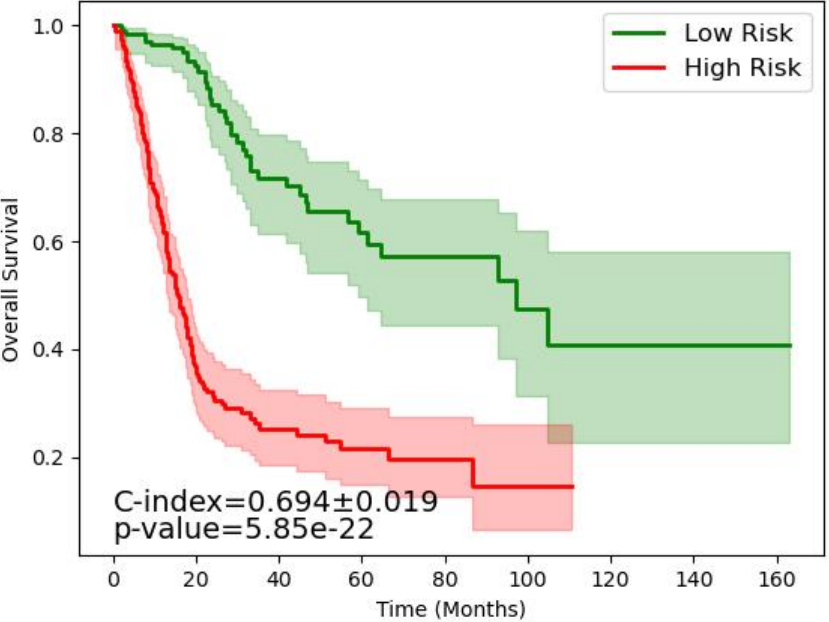}
        \caption{BLCA dataset}
    \end{subfigure}
    \hfill
    \begin{subfigure}{0.24\linewidth}
		\centering
		\includegraphics[width=0.95\linewidth]{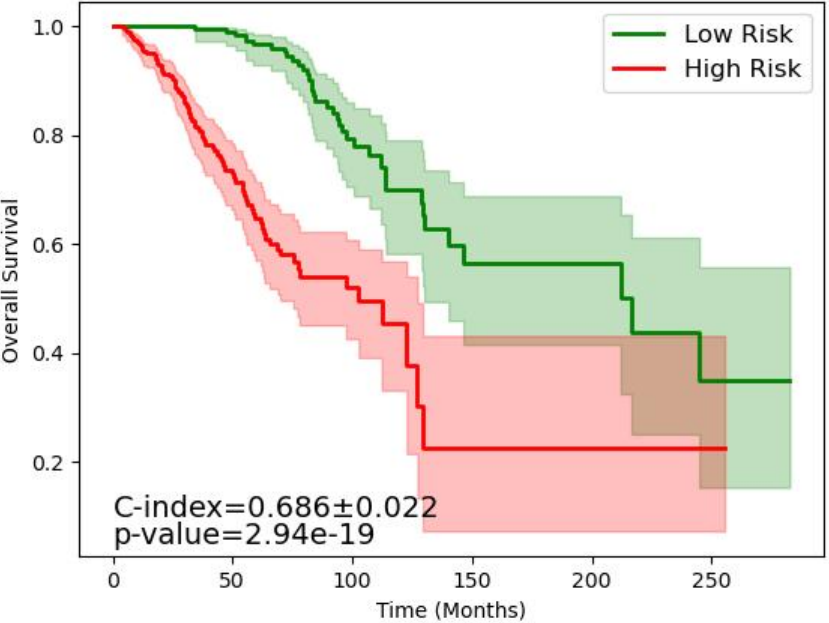}
		\caption{BRCA dataset}
	\end{subfigure}
    \hfill
	\begin{subfigure}{0.24\linewidth}
		\centering
		\includegraphics[width=0.95\linewidth]{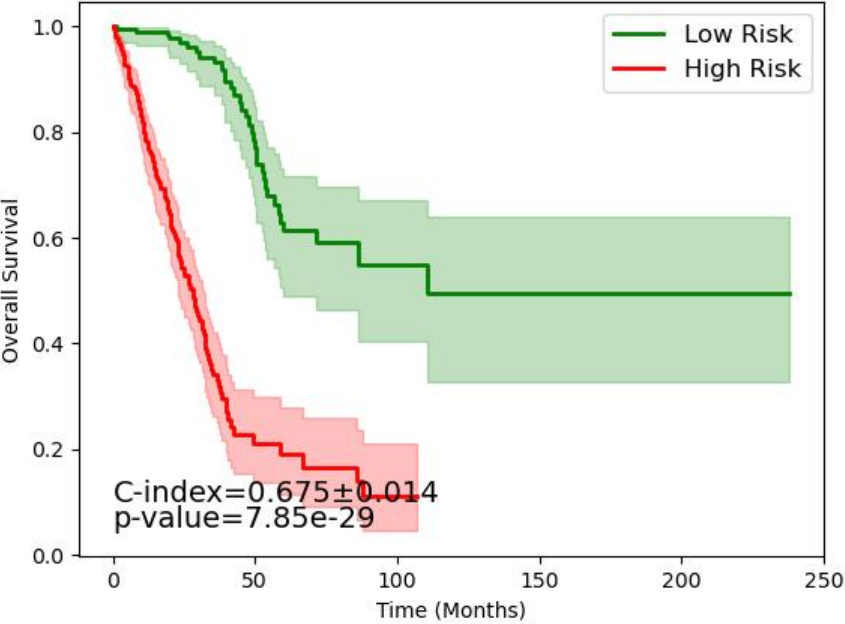}
		\caption{LUAD dataset}
	\end{subfigure}
       \caption{Visualization of Kaplan-Meier Analysis, where patient stratifications of low risk (green) and high risk (red) are presented. Shaded areas refer to the confidence intervals.
       }
    \label{fig:km}
        \vspace{-2ex}
\end{figure*}

\begin{figure*}[t]
    \centering
    \begin{subfigure}{0.24\textwidth}
        \centering
        \includegraphics[width=0.95\textwidth]{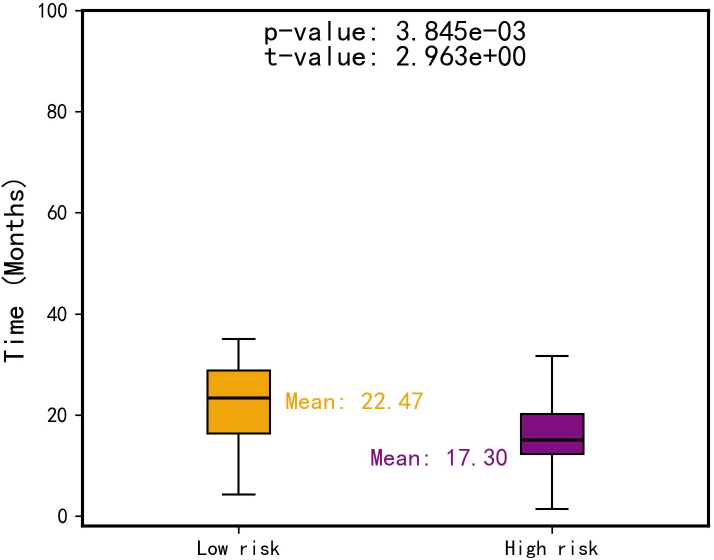}
        \caption{LC dataset}
    \end{subfigure}
    \hfill
    \begin{subfigure}{0.24\textwidth}
        \centering
        \includegraphics[width=0.95\textwidth]{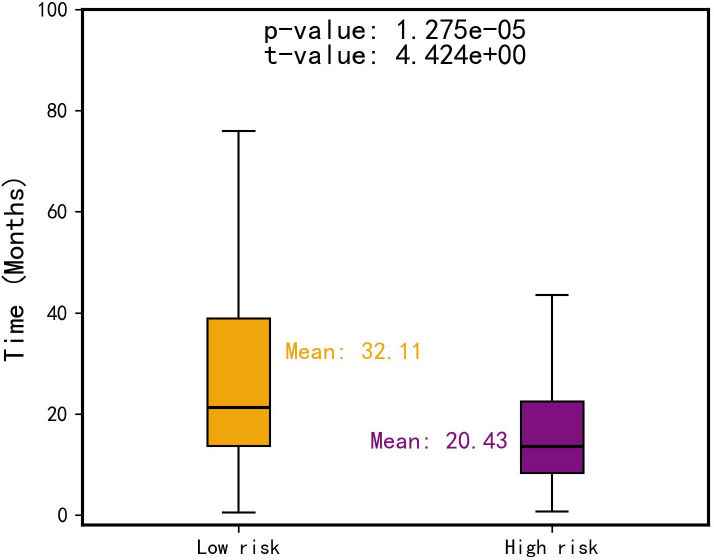}
        \caption{BLCA dataset}
    \end{subfigure}
    \hfill
    \begin{subfigure}{0.24\linewidth}
		\centering
		\includegraphics[width=0.95\linewidth]{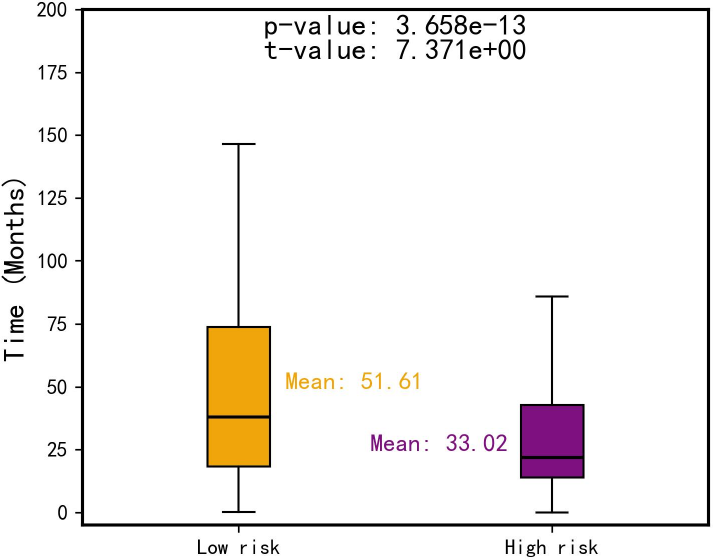}
		\caption{BRCA dataset}
	\end{subfigure}
    \hfill
	\begin{subfigure}{0.24\linewidth}
		\centering
		\includegraphics[width=0.95\linewidth]{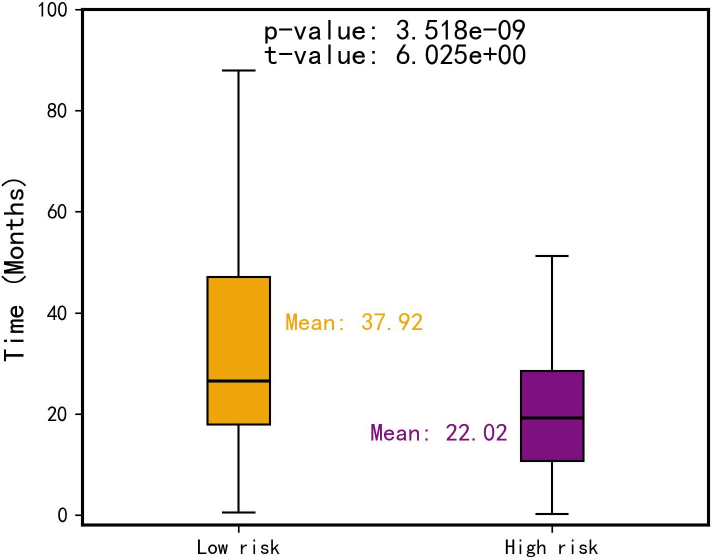}
		\caption{LUAD dataset}
	\end{subfigure}
       \caption{Visualization of T-test Analysis, where patient box-plots of low risk (orange) and high risk (purple) are presented.
       }
    \label{fig:t_test}
        \vspace{-2ex}
\end{figure*}

\begin{table}[t]
\caption{The impact of different fine-grained token lengths (TL) in two-level MoE modules on four datasets.}
\centering
\renewcommand{\arraystretch}{0.85} 
    \begin{tabular}{l|cccc}
    \toprule
        \multirow{2}{*}{\bfseries TL}
        &\multicolumn{4}{c}{\bfseries Dataset} \\
     &LC &BLCA &BRCA &LUAD \\
     \midrule
     2   & 0.678$_{\pm0.022}$ & 0.651$_{\pm0.020}$ & 0.614$_{\pm0.058}$ & 0.634$_{\pm0.021}$ \\
     8   & 0.657$_{\pm0.026}$ & 0.652$_{\pm0.022}$ & 0.643$_{\pm0.011}$ & 0.662$_{\pm0.026}$ \\
     16  & 0.679$_{\pm0.029}$ & 0.644$_{\pm0.026}$ & 0.630$_{\pm0.022}$ & 0.669$_{\pm0.020}$ \\
     32  &\textcolor{blue}{\textit{0.681}}$_{\pm0.019}$ & \textcolor{blue}{\textit{0.690}}$_{\pm0.024}$ & \textcolor{blue}{\textit{0.682}}$_{\pm0.047}$ & \textcolor{red}{\underline{0.677}}$_{\pm0.038}$ \\
     64  & \textcolor{red}{\underline{0.683}}$_{\pm0.029}$ 
         & \textcolor{red}{\underline{0.694}}$_{\pm0.019}$ 
         & \textcolor{red}{\underline{0.686}}$_{\pm0.022}$ 
         & \textcolor{blue}{\textit{0.675}}$_{\pm0.014}$ \\
     128 & 0.678$_{\pm0.025}$ & 0.672$_{\pm0.017}$ & 0.657$_{\pm0.029}$ & 0.671$_{\pm0.043}$ \\
     256 & 0.667$_{\pm0.043}$ & 0.659$_{\pm0.040}$ & 0.644$_{\pm0.040}$ & 0.668$_{\pm0.040}$ \\
    \bottomrule
    \end{tabular}
    \label{tab:token_length}
\end{table}
\begin{figure*}[h]
    \centering
    \begin{subfigure}[t]{0.49\textwidth}
        \centering
        \includegraphics[width=\textwidth]{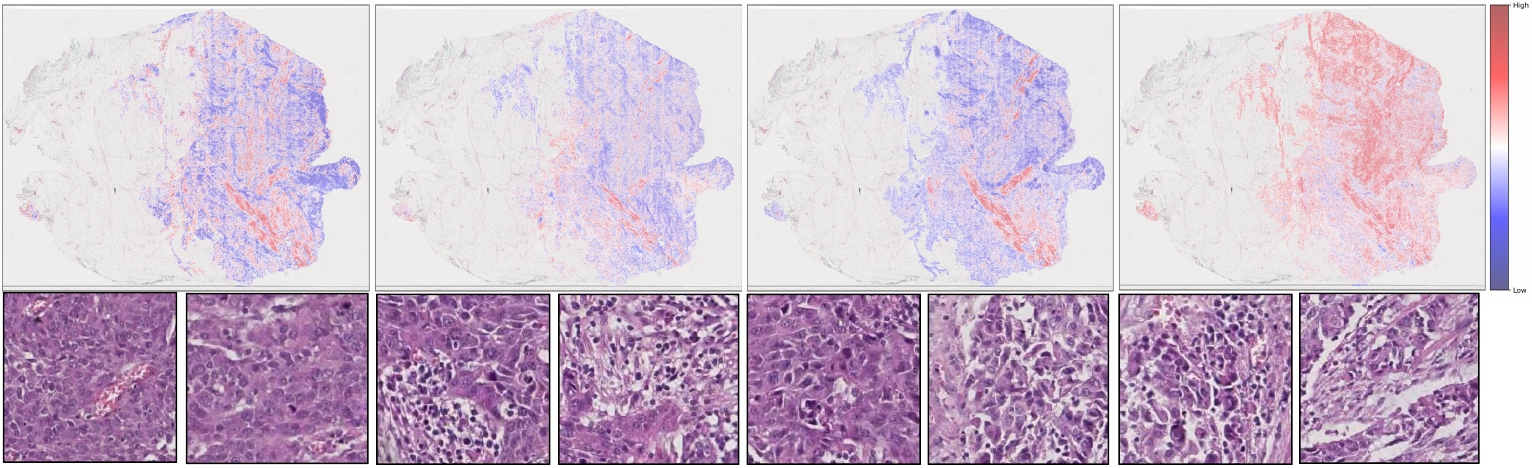}
        \caption{Intra-Pathology Attention Map from Expert 2, 4, 6, 7}
    \end{subfigure}
    \hfill
    \begin{subfigure}[t]{0.49\textwidth}
        \centering
        \includegraphics[width=\textwidth]{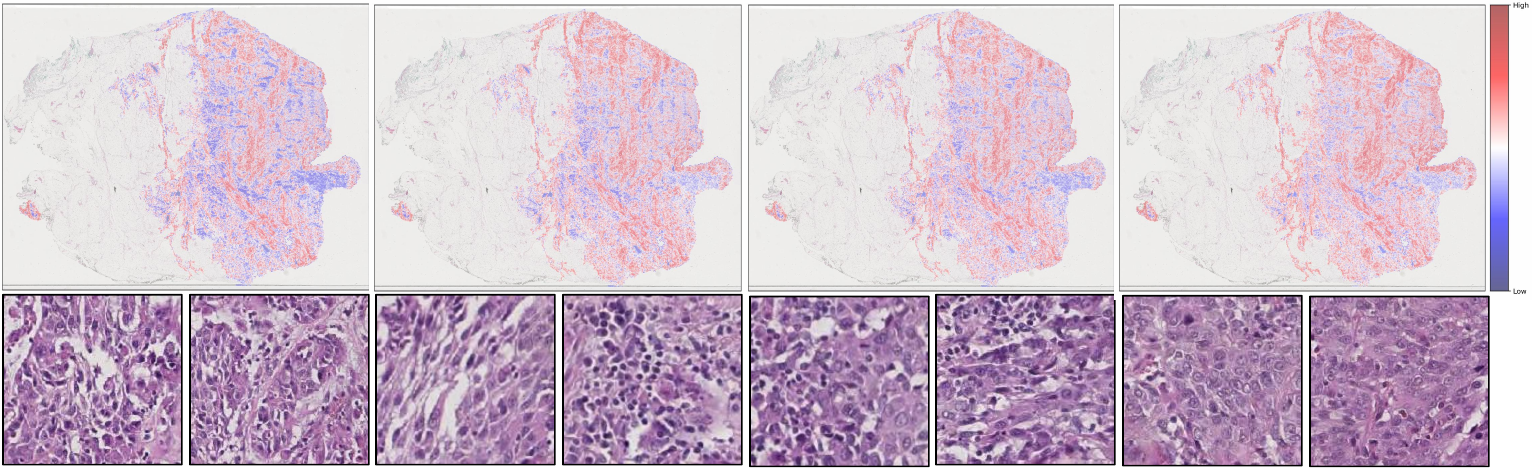}
        \caption{Inter-Modality Attention Map from Expert 0, 1, 5, 6}
    \end{subfigure}
    \caption{Visualization experiment on a TCGA-BLCA sample. In each sub-figure, The left part shows the fine-grained feature attention heatmap output by four First-level routed pathology experts, and the right part shows the fine-grained feature attention heatmap output by four Second-level experts.}
    \label{fig:viss}
    \vspace{-2ex}
\end{figure*}

\begin{figure}[t]
    \centering
    \begin{subfigure}[t]{0.29\textwidth}
        \centering
        \includegraphics[width=\textwidth]{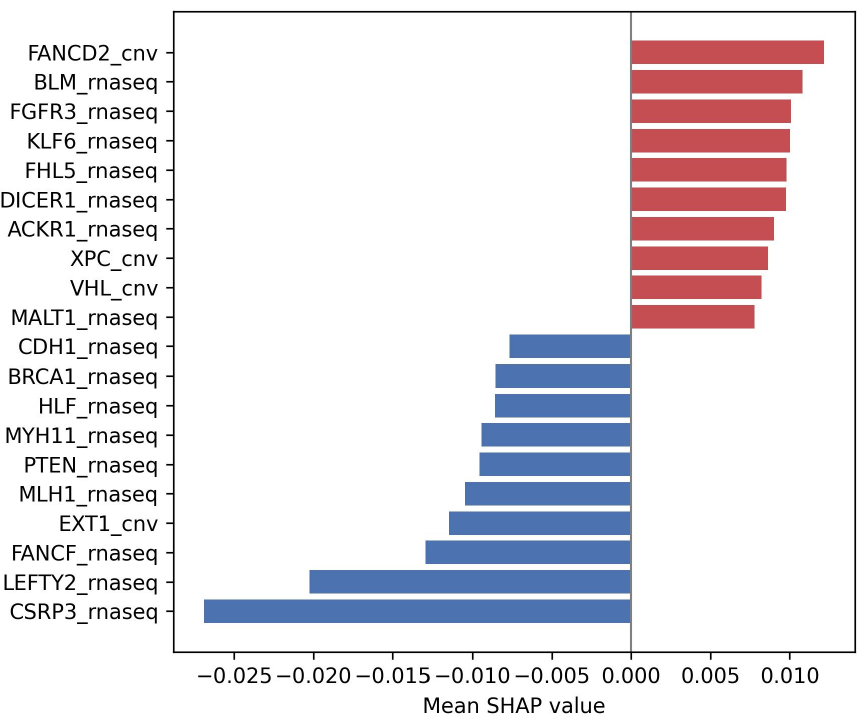}
    \end{subfigure}

    \caption{SHAP analysis of geonmic features on a TCGA-BLCA sample.}
    \label{fig:shap}
    \vspace{-2ex}
\end{figure}

\noindent \textbf{Settings of Framework Components.} We conduct experiments to evaluate the effectiveness of the framework components. We accomplish the fine-grained and de-redundancy ablation experiments by not dividing the features $V_1,V_2$ into multiple tokens and by substituting $V_1,V_2$ with the purified features output by the first-level shared experts. Results are shown in Tab.~\ref{tab:ablation}. We observe that HDMoE achieves the best performance on all datasets when all five components are used. In contrast, removing any component leads to performance degradation. For instance, the lack of fine-grained and de-redundancy components causes 2.8\%, 2.4\% performance degradation on BLCA dataset, which highlights the importance of exploring inter-modality fine-grained feature relationships and removing feature redundancies before decoupling. 
Furthermore, omitting the decoupling term leads to a loss of modality decoupling, resulting in modality confusion in the second-level MoE and negatively impacting model performance.

\noindent \textbf{Impact of Distance Metrics.} We investigate the effect of different distance metrics used in decoupling loss. By default, we employ Cosine Similarity to measure the distance between decoupled feature representations. Furthermore, we compared Cosine Similarity with alternative metrics, including L1 norm, Kullback-Leibler divergence, and Mean Squared Error. As shown in the right part of Tab.~\ref{tab:dis}, using Cosine Similarity metric empirically produces better results.

\noindent \textbf{Impact of Different Expert Numbers.} Generally, the number of experts in MoE module plays a crucial role in model performance. Therefore, we investigate the impact of different expert numbers on experimental results, as shown in Fig.~\ref{fig:num_expert}. On LC, BLCA, and BRCA datasets, the optimal performance is achieved with 8 experts, yielding results of 0.683, 0.694, and 0.686, respectively. Thus, we set the number of experts $N=8$. An insufficient number of experts may limit the model's ability to capture complex patterns, while an excessive number may lead to insufficient training for some experts due to the limited dataset size, preventing effective learning of meaningful representations.

\noindent \textbf{Impact of Different Token Lengths.} We analyze the impact of different fine-grained token lengths in the two-level MoE modules. As shown in Tab.~\ref{tab:token_length}, lengths of 64 and 32 consistently yield the best and second-best performance across the four datasets. We attribute this to an optimal balance between the shared expert's functional role and the preservation of contextual integrity. Shorter tokens suffer from insufficient semantic content, hindering meaningful expert specialization. In contrast, longer tokens reduce the number of available tokens for routing, limiting the shared expert's ability to extract common patterns. This reduces its ability to capture generic knowledge and reduce redundancy. Therefore, moderate token lengths (32 or 64) strike a favorable trade-off by ensuring semantic content integrity while enabling the shared expert to effectively learn shared representations.

\noindent \textbf{Impact of Modality Importance.} We conduct unimodal ablation experiments to validate modality importance, as shown in Tab.~\ref{tab:modality_performance}. The results suggest that the importance of Genomic or MRI modality surpasses that of Pathology modality. Taking the Genomic and Pathology modalities as an example, the precise information provided by genomic data—such as mutations, gene expression, and other biomarkers—is generally more robust than image-based pathological features, which may be influenced by factors such as tissue preparation quality and staining variations.

\noindent \textbf{Impact of Shared Expert.} To verify that shared experts can capture general knowledge across different fine-grained tokens and reduce intra-modality redundancy, we conduct Feature de-redundancy ablation experiments, as illustrated in Fig.~\ref{fig:der}. Using the BLCA dataset as an example, we compare the average self-correlation heatmaps of different modality tokens in the First-Level MoE before and after shared experts. The results show that the total non-diagonal correlations of pathological and genomic tokens decreased by 0.28 and 0.46, indicating that shared experts reduce redundancy among intra-modality tokens. More Results are shown in Appendix~\ref{ap:exp}.

\noindent \textbf{Allocations of Routed Experts.} To validate that different experts possess distinct feature analysis capabilities, we analyze the expert allocation patterns of the model across the four datasets, as shown in Fig.~\ref{fig:exp_routed}. It is evident that the frequency of allocated experts varies among different expert modules. For instance, in BRCA dataset, expert 1 in pathology experts is invoked most frequently, highlighting its importance, whereas experts 0 and 4 are invoked less, suggesting their weaker significance.

\noindent \textbf{Robustness Evaluation of RFR.} We verify the stability of the proposed RFR module by repeatedly evaluating the same trained model on the identical test dataset. The results and analysis are reported in Appendix~\ref{ap:rob}.

\noindent \textbf{Impact of Balance Factors $\alpha$ and $\beta$.} \re{We evaluate the sensitivity of the balance factors $\alpha$ and $\beta$ in Eq.~(\ref{eq:totalloss}) by varying their values and examining the corresponding model performance, as shown in Appendix~\ref{ap:hyper}. The results indicate that the values of $\alpha$ and $\beta$ have minimal impact on the model (C-index fluctuation range $<$0.03).}

\subsection{Visualization Analysis}
\noindent\textbf{Kaplan-Meier Analysis.} The Kaplan-Meier method~\cite{kaplan1958nonparametric} is a nonparametric approach for estimating survival functions and analyzing time-to-event data. Based on the median survival time, we stratify patients into high- and low-risk groups, depicted as red and green curves. The log-rank test evaluates the statistical significance of differences between the two groups, with a lower p-value indicating better stratification. As shown in Fig.~\ref{fig:km}, our framework yields p-values below 0.05 on all datasets, confirming a significant distinction between the high- and low-risk groups.

\noindent\textbf{T-test Analysis.} The T-test is a statistical hypothesis testing method used to assess whether there exists a significant difference between the means of two groups. In this analysis, the t-value reflects the magnitude of the difference between group means relative to their variability, while the p-value represents the probability of observing such results under the null hypothesis. Using median survival time, we divide patients into high- and low-risk groups for T-test evaluation. As shown in Fig~\ref{fig:t_test}, our framework achieves favorable t-values and p-values on all datasets.

\noindent\textbf{Attention Analysis of Decoupled Features.} 
We visualize the attention regions of the output feature tokens from the top four routed experts on WSI, using the TCGA-2F-A9KO sample as an example (Fig.~\ref{fig:viss}). The highlighted patches correspond to areas each expert attends to, revealing histopathologically meaningful structures. These include cellular and nuclear pleomorphism, increased nuclear-to-cytoplasmic ratio, coarse chromatin, prominent nucleoli, and abnormal mitotic figures—features commonly associated with malignancy~\cite{ohashi2018prognostic,KADOTA2012260}. This visualization confirms that the model focuses on diagnostically relevant regions, and that each expert contributes by encoding distinct aspects of the input tokens.

\noindent\textbf{SHAP Analysis of Genomic Features.} \re{We further conduct SHAP analysis on genomic features for the same TCGA-2F-A9KO sample, as shown in Fig.~\ref{fig:shap}. The results show that several high-contribution genes, including FGFR and CDH1. As these genes have been reported to be associated with prognosis~\cite{hernandez2006prospective,yang2020reduced}, this result suggests that the model can capture biologically relevant molecular patterns and provides supportive evidence for its potential in scientific discovery.

}
\section{Discussion \& Conclusion}
In this paper, we proposed a HDMoE framework for multimodal cancer survival analysis. The first-level MoE is able to reduce intra-modality redundancies and extract token-level specific features within each modality. The second-level MoE further captures complementary information between modalities and obtains inter-modality shared and specific features. Besides, RFR module was introduced to enrich inter-modality relationships while achieving feature fusion. 
\re{Extensive experiments on four datasets validate the effectiveness and clinical impact of the proposed framework. Quantitatively, HDMoE achieves the highest C-index of 0.683 on LC dataset and 0.694, 0.686, and 0.675 on BLCA, BRCA, and LUAD datasets, respectively—representing an average improvement of 1.5\% over the best existing multimodal baselines. In addition, Kaplan–Meier survival analyses demonstrate a clear separation between high- and low-risk groups with p-values $<$ 0.05 on all datasets, and T-test hypothesis confirm statistically significant distinctions in predicted risk distributions. Beyond predictive performance, the visualization analyses indicate that the model can identify prognostically relevant patterns, including clinically meaningful pathological regions and molecular features associated with survival.
These results suggest that HDMoE can learn effective multimodal representations for risk stratification by better handling modality heterogeneity, redundancy, and cross-modal interactions.
Collectively, the framework improves predictive accuracy while also providing supportive evidence that it can capture survival-relevant multimodal signals.}

\section{Limitations \& Ethical Considerations}
In this paper, several limitations should be acknowledged. First, the framework assumes the availability of paired multimodal data, which may limit its applicability in scenarios with missing or incomplete modalities. Future research could explore modality-agnostic learning or cross-modal generation to address this challenge. Second, due to the current limitation of dataset scale, multi-center validation could not be conducted in this study. In future work, we plan to collect and integrate large-scale, multi-center datasets to further assess the generalizability of the proposed framework.

We confirm that we have read the conference’s position on issues involved in ethical publication and affirm that this paper is consistent with those guidelines. Ethical approval for the collection and use of the private LC dataset was obtained from the Ethics Committee of Sun Yat-sen University Cancer Center under protocol number B2021-214-Y03.

\begin{acks}
This research was partially supported by the National Natural Science Foundation of China under Grant No. 92259202 and No. 62476246, "Pioneer" and "Leading Goose" R\&D Program of Zhejiang under Grant No. 2025C02120, and GuangZhou City’s Key R\&D Program of China under Grant No. 2024B01J1301. 
\end{acks}

\bibliographystyle{ACM-Reference-Format}
\bibliography{sample-base}

\clearpage
\appendix

\section{Appendix}

\subsection{Related Work}
\subsubsection{Multimodal Survival Prediction}
Survival analysis aims to study the relationship between time-to-event outcomes (\eg patient survival time) and covariates (\eg clinical features, WSIs, and Genomic Profiles). 
In recent years, multimodal learning~\cite{CMTAs,MCAT,yang2024facilitating} has emerged as a key direction to improve survival prediction performance by integrating complementary information from various modalities to overcome the limitations of unimodal approaches.
For example, Chen et al.~\cite{chen2020pathomic} proposed a multimodal feature interaction method based on the Kronecker product, combined with a gated attention mechanism to control the representation weights of each modality. 
Studies~\cite{li2022hfbsurv,CMTAs,MCAT} employed attention-based fusion strategies to enhance inter-modality interactions.
Xiong et al.~\cite{xiong2024mome} introduced a progressive fusion approach based on MoE, where each expert performs different feature interactions like concatenation, averaging, and attention mechanism.
Other studies~\cite{CFDL,zhou2024cohort,PIBD,wang2025decouple} introduced a decoupling-fusion paradigm, decomposing multimodal features into modality-specific and modality-shared components before fusion.
Building upon the decoupling-fusion paradigm, our work proposes a hierarchical purify-then-decouple fusion framework to mitigate noise and redundancy in raw multimodal features, which often interfere with decoupling and cross-modal modeling. Our framework ultimately achieves efficient and robust multimodal survival prediction.

\subsubsection{Mixture-of-Experts}

Mixture of Experts (MoE) was first proposed by Jacobs et al.~\cite{Jacobs1991AdaptiveMO}, as an ensemble framework comprising multiple sub-networks referred to as ``expert''.
Subsequently, Eigen et al.~\cite{eigen2013learning} embedded MoE into the neural network layer, where a learnable gating network activates all experts and assigns the corresponding weights to them for weighted aggregation.
According to the activation mechanism, MoE can be categorized into Dense MoE and Sparse MoE (SMoE). Early MoE studies~\cite{ma2018modeling,yuksel2012twenty,chen1999improved} 
tended to densely activate all experts, leading to high computational costs~\cite{eigen2013learning}. To address this issue, many studies~\cite{shazeer2017outrageously,lepikhin2020gshard,fedus2022switch,gross2017hard,Shao2024DeepSeekV2AS} adopt a Top-K routing mechanism. For different fine-grained tokens, only experts corresponding to the Top-K weight are activated, thus greatly reducing the computational overhead.
For example, Fedus et al.~\cite{fedus2022switch} applied SMoE to the feed-forward network layers of the transformer using a Top-1 routing strategy, improving the efficiency and scalability of the model.
Given that SMoE not only offers a lightweight architectural design but also enhances the model's ability to learn fine-grained contextual representations through flexible expert allocation, we adopt it as the core component of our model. Furthermore, by integrating a shared expert~\cite{Shao2024DeepSeekV2AS} and a decoupling loss function, we endow our \modelname with the capability of feature purification and decoupling, thereby improving performance in survival analysis task.

\subsection{Survival Loss and C-Index}
\label{ap:loss}
\subsubsection{Survival Loss}
Following previous work~\cite{zhou2024cohort}, we convert survival time prediction from a continuous-time regression problem into a discrete-time, period-based classification problem. Specifically, given the right-censored patient-level survival time in months $T_{\text{cont}} \in [0,\infty)$, we partition the timeline into $K$ non-overlapping bins $[t_0,t_1), [t_1,t_2), \dots, [t_{K-1}, t_K)$, where $t_0=0$, $t_K=\infty$, and ${t_1,\dots,t_{K-1}}$ are determined by the quantiles (e.g., quartiles) of event times from uncensored patients. For each patient $i$ with follow-up time $T_{i,\text{cont}}$, we assign a discrete time-period label $n_i \in {1,\dots,K}$ as the index of the bin containing $T_{i,\text{cont}}$. Here, $K=4$.

For each sample, we define ${h_{\text{output}}, c, n}$, where $h_{\text{output}}={h_1,\dots,h_K}$ denotes the predicted hazard probabilities over time bins, $n$ is the ground-truth discrete label, and $c\in{0,1}$ indicates the censoring status. The hazard can be interpreted as the conditional event probability at period $j$:
\begin{equation}
h_j \;\triangleq\; f_{\text{hazard}}(j)
\;=\; \mathbb{P}(T=j \mid T\ge j, \cdot),
\end{equation}
which relates to the discrete survival function by
\begin{equation}
f_{\text{surv}}(h_{\text{output}}, n)
= \mathbb{P}(T>n \mid \cdot)
= \prod_{j=1}^{n}(1-h_j).
\end{equation}

Based on this, the likelihood for an uncensored patient ($c=0$) with event occurring in period $n$ is:
\begin{equation}
\mathbb{P}(T=n)= h_n \cdot \prod_{j=1}^{n-1}(1-h_j),
\end{equation}
while for a censored patient ($c=1$) whose follow-up ends in period $n$, the likelihood is
\begin{equation}
\mathbb{P}(T>n)= \prod_{j=1}^{n}(1-h_j).
\end{equation}

Therefore, we adopt the negative log-likelihood (NLL) with censoring to supervise survival prediction:
\begin{equation}
\begin{aligned}
\mathcal{L}_{\text {surv }}= & -c \log \left(f_{\text {surv }}(h_{\text{output}}, n)\right) \\
&-(1-c) \log \left(h_n\right)
-(1-c) \log \left(f_{\text {surv}}(h_{\text{output}}, n-1)\right),
\end{aligned}
\label{eq:loss_sur}
\end{equation}
where $f_{\text{surv}}(h_{\text{output}}, n-1)=\prod_{j=1}^{n-1}(1-h_j)$. Intuitively, the loss encourages (i) a high survival probability beyond the censoring interval for censored cases, and (ii) a high survival probability before period $n$ together with a high hazard at period $n$ for uncensored cases.
\subsubsection{C-Index}
The Concordance Index (C-index)~\cite{harrell1982evaluating} is a widely used metric for evaluating survival prediction models. It measures the agreement between the predicted risk scores and the observed survival outcomes by assessing whether patients with higher predicted risk tend to experience events earlier. Formally, the C-index is defined as the proportion of concordant pairs among all comparable patient pairs. A pair of patients $(i,j)$ is considered comparable if the patient with the shorter observed time experienced an event (i.e., if $t_i < t_j$, then $\delta_i = 1$), where $t$ denotes the observed time and $\delta$ indicates whether the event is observed. The pair $(i,j)$ is concordant if the predicted risk score is higher for the patient with the shorter event time, i.e., $h_i > h_j$ when $t_i < t_j$; otherwise it is discordant~\cite{JMLRv21}.

\subsection{Additional Experiment Results}
\subsubsection{Impact of Shared Expert}
\label{ap:exp}
\begin{figure}[t]
    \centering
    \begin{subfigure}[t]{0.49\textwidth}
        \centering
        \includegraphics[width=\textwidth]{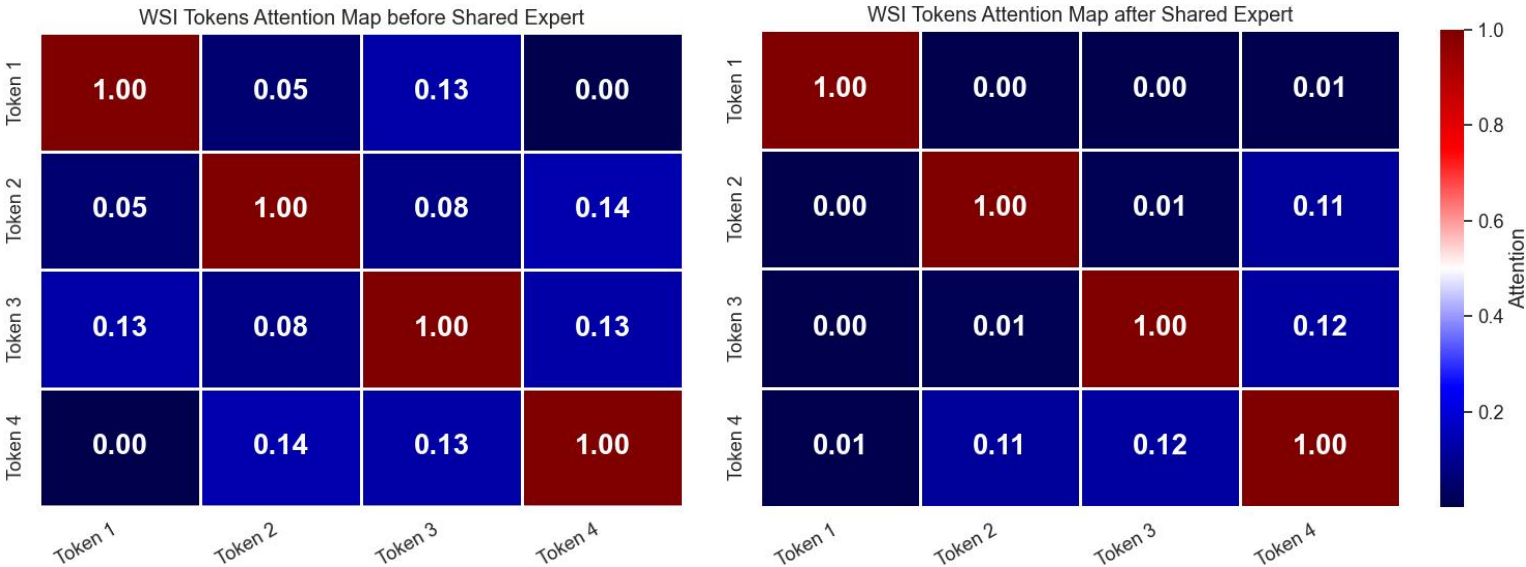}
        \caption{Pathology token correlations before and after shared expert}
    \end{subfigure}
    \hfill
    \begin{subfigure}[t]{0.49\textwidth}
        \centering
        \includegraphics[width=\textwidth]{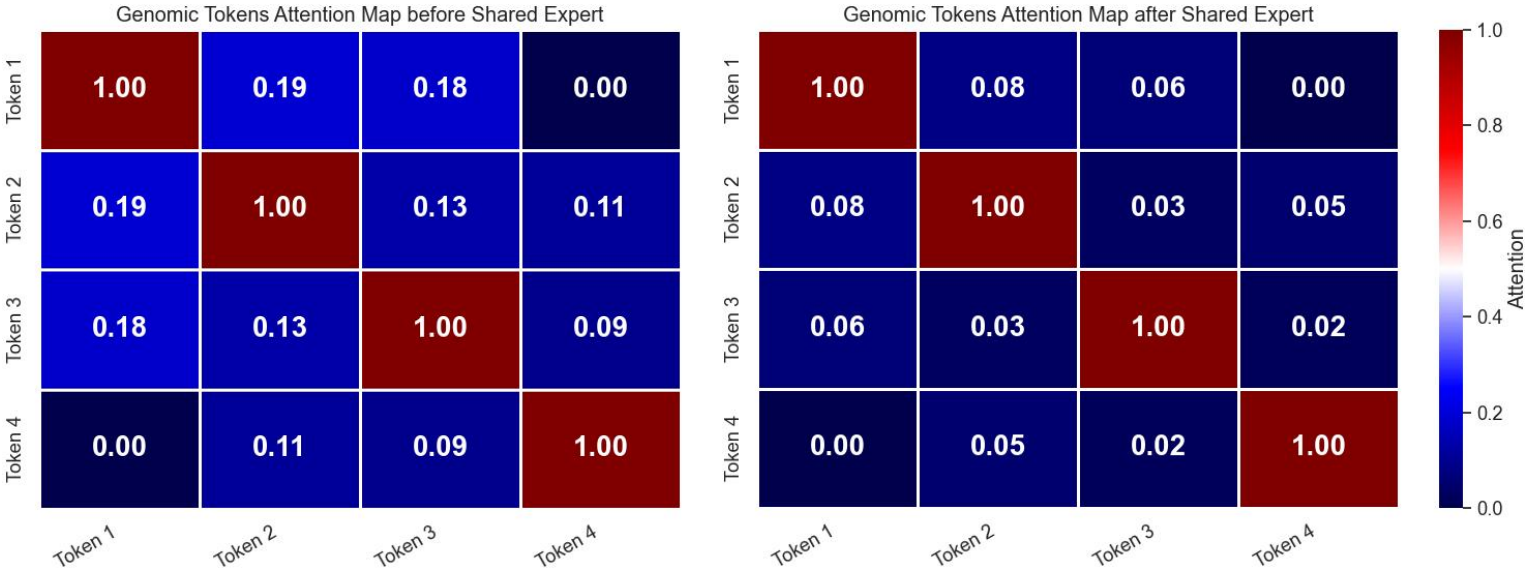}
        \caption{Genomic token correlations before and after shared expert}
    \end{subfigure}
    \caption{Feature de-redundancy experiments on LC dataset. Each sub-figure shows average correlation heatmaps of tokens before and after shared expert.}
    \label{fig:der2}
    \vspace{-2ex}
\end{figure}
\begin{figure}[t]
    \centering
    \begin{subfigure}[t]{0.49\textwidth}
        \centering
        \includegraphics[width=\textwidth]{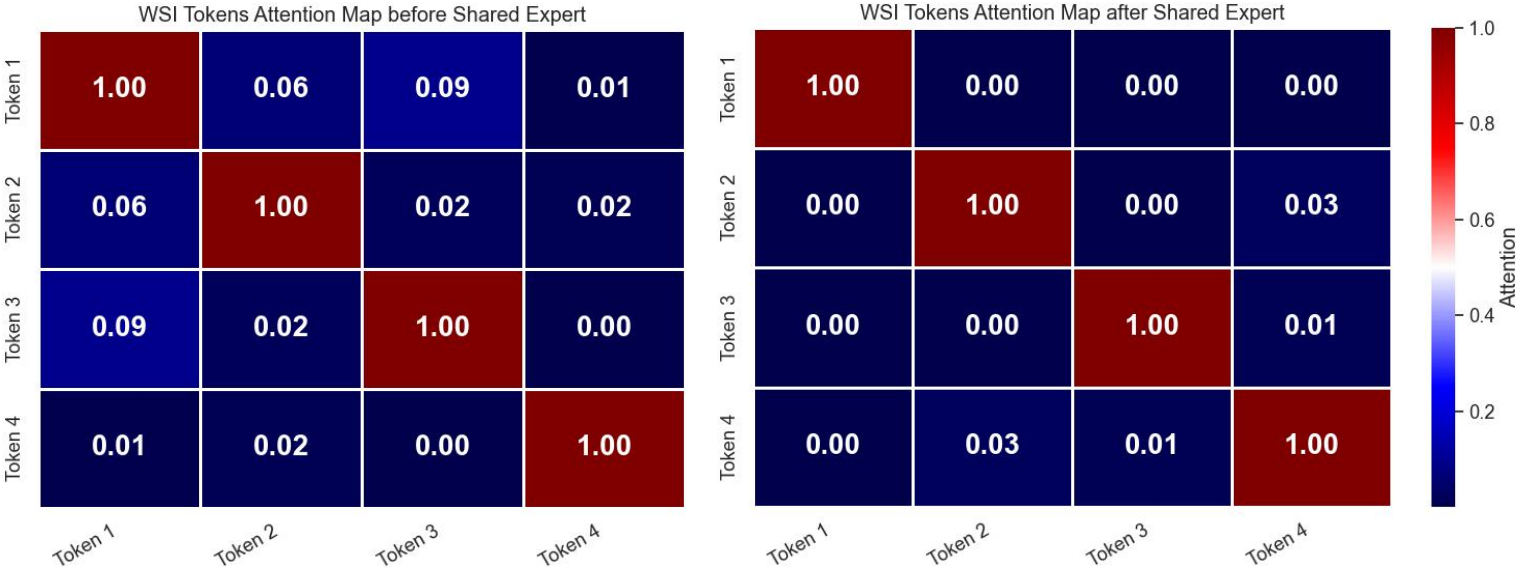}
        \caption{Pathology token correlations before and after shared expert}
    \end{subfigure}
    \hfill
    \begin{subfigure}[t]{0.49\textwidth}
        \centering
        \includegraphics[width=\textwidth]{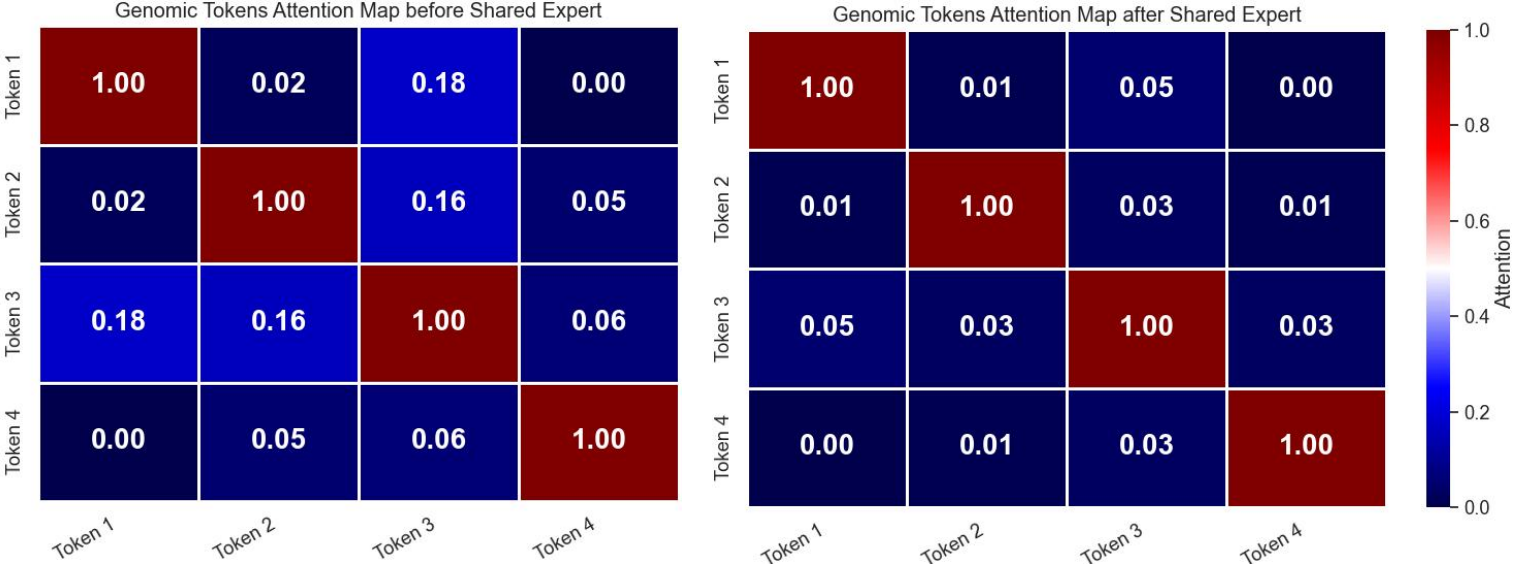}
        \caption{Genomic token correlations before and after shared expert}
    \end{subfigure}
    \caption{Feature de-redundancy experiments on TCGA-BRCA dataset. Each sub-figure shows average correlation heatmaps of tokens before and after shared expert.}
    \label{fig:der3}
    \vspace{-2ex}
\end{figure}
\begin{figure}[t]
    \centering
    \begin{subfigure}[t]{0.49\textwidth}
        \centering
        \includegraphics[width=\textwidth]{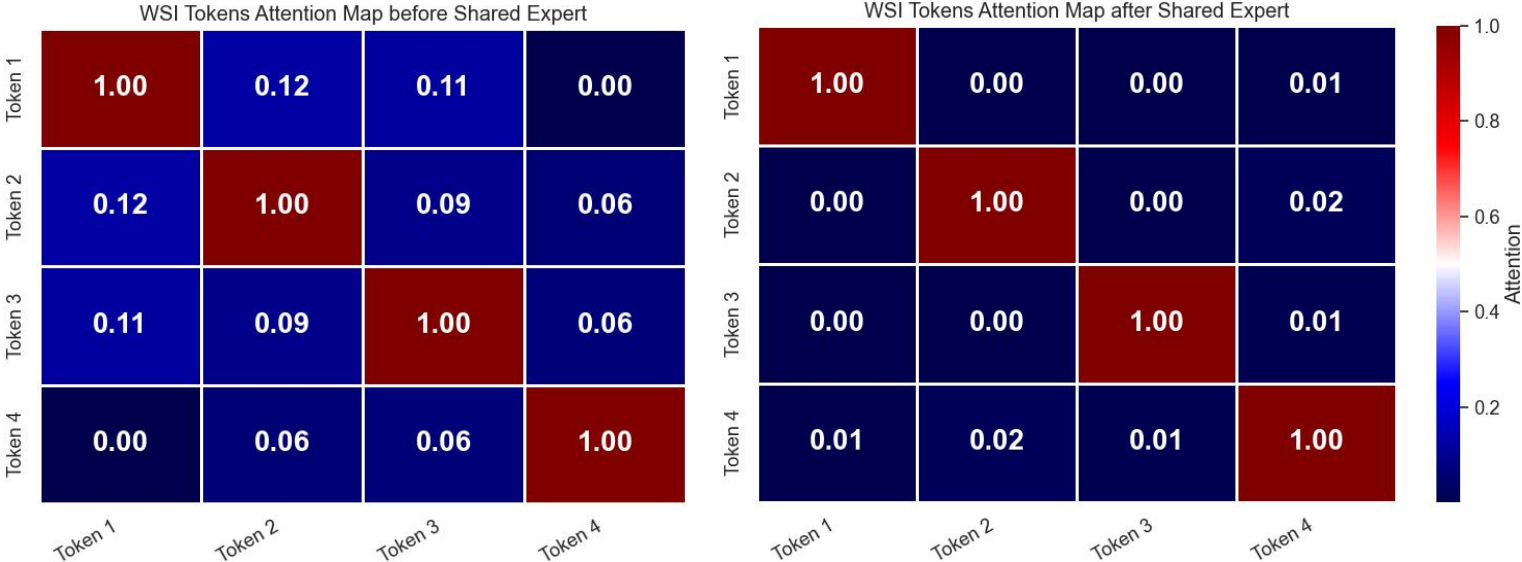}
        \caption{Pathology token correlations before and after shared expert}
    \end{subfigure}
    \hfill
    \begin{subfigure}[t]{0.49\textwidth}
        \centering
        \includegraphics[width=\textwidth]{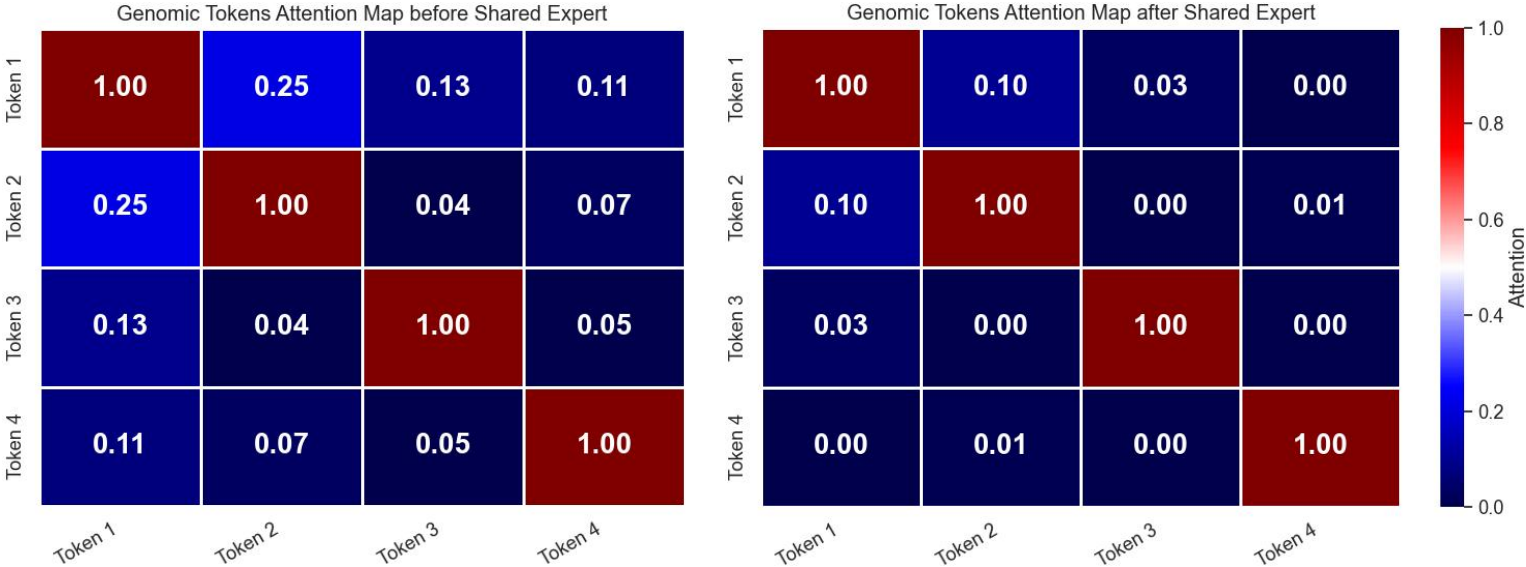}
        \caption{Genomic token correlations before and after shared expert}
    \end{subfigure}
    \caption{Feature de-redundancy experiments on TCGA-LUAD dataset. Each sub-figure shows average correlation heatmaps of tokens before and after shared expert.}
    \label{fig:der4}
    \vspace{-2ex}
\end{figure}

More results about the impact of shared expert are shown in Figure~\ref{fig:der2}, Figure~\ref{fig:der3} and Figure~\ref{fig:der4}, where we summarize the statistics on LC, TCGA-BRCA, and TCGA-LUAD. Across all these datasets, the total average non-diagonal self-correlations of both pathology and genomics tokens consistently decrease after shared experts, confirming that shared experts effectively mitigate intra-modality redundancy.

\subsubsection{Robustness Evaluation of RFR} 
\label{ap:rob}
To evaluate the test-time stability of the Random Feature Reorganization (RFR) algorithm, we run the same trained model on the identical test dataset for five repeated trials. As shown in Tab.~\ref{tab:multi_test}, the predictions are nearly identical across runs. These results indicate that RFR not only makes the model invariant to different feature arrangements, but also enhances its ability to capture diverse feature relationships.
\begin{table}[t]
    \centering
    \caption{Repeated-test results (mean $\pm$ std) on four datasets. The `Average' row reports the mean and standard deviation of the performance means from three repeated tests.}
    \begin{tabular}{l|cccc}
        \toprule
        Rounds & LC & BLCA & BRCA & LUAD \\
        \midrule
        1 & 0.679$_{\pm\text{0.027}}$ & 0.688$_{\pm\text{0.024}}$ & \textcolor{blue}{\textit{0.689}}$_{\pm\text{0.022}}$ & 0.670$_{\pm\text{0.024}}$ \\
        2 & \textcolor{blue}{\textit{0.684}}$_{\pm\text{0.029}}$ & \textcolor{red}{\underline{0.697}}$_{\pm\text{0.022}}$ & 0.683$_{\pm\text{0.027}}$ & 0.673$_{\pm\text{0.020}}$ \\
        3 & 0.677$_{\pm\text{0.030}}$ & \textcolor{blue}{\textit{0.693}}$_{\pm\text{0.017}}$ & \textcolor{red}{\underline{0.691}}$_{\pm\text{0.018}}$ & \textcolor{red}{\underline{0.679}}$_{\pm\text{0.028}}$ \\
        4 & 0.681$_{\pm\text{0.024}}$ & 0.691$_{\pm\text{0.028}}$ & 0.683$_{\pm\text{0.023}}$ & 0.672$_{\pm\text{0.019}}$ \\
        5 & \textcolor{red}{\underline{0.688}}$_{\pm\text{0.018}}$ & 0.689$_{\pm\text{0.031}}$ & 0.679$_{\pm\text{0.019}}$ & \textcolor{blue}{\textit{0.677}}$_{\pm\text{0.027}}$ \\
        \midrule
        Average & 0.682$_{\pm\text{0.004}}$ & 0.692$_{\pm\text{0.003}}$ & 0.685$_{\pm\text{0.004}}$ & 0.674$_{\pm\text{0.003}}$ \\
        \bottomrule
    \end{tabular}
    \label{tab:multi_test}
\end{table}

\subsubsection{Impact of Balance Factors $\alpha$ and $\beta$.} 
\label{ap:hyper}
\begin{figure}[t]
    \centering
    \begin{subfigure}[t]{0.49\textwidth}
        \centering
        \includegraphics[width=\textwidth]{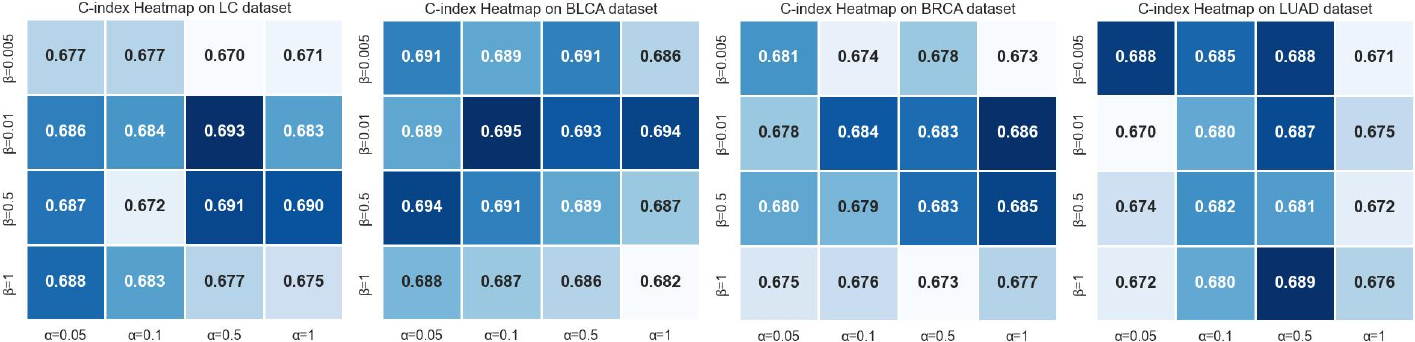}
    \end{subfigure}
    \caption{Sensitivity analysis on the balance factors $\alpha$ and $\beta$}
    \label{fig:grid}
    \vspace{-2ex}
\end{figure}

To validate that the proposed framework is not overly sensitive to a narrow range of hyperparameter values, we perform a sensitivity analysis on the balance factors $\alpha$ and $\beta$ in Eq.~(\ref{eq:totalloss}). The results are shown in Figure~\ref{fig:grid}. It shows that the values of $\alpha$ and $\beta$ have minimal impact on the model (C-index fluctuation range <0.03).

\end{document}